\documentclass[runningheads]{llncs}

\usepackage{eccv}

\usepackage{eccvabbrv}

\usepackage{graphicx}
\usepackage{booktabs}
\usepackage[normalem]{ulem}  

\usepackage[accsupp]{axessibility}  

\usepackage{hyperref}

\usepackage{orcidlink}
\usepackage{enumitem}
\usepackage{comment}
\usepackage{bm}
\usepackage{algpseudocode}
\usepackage{algorithm}
\usepackage{multirow}
\usepackage{marvosym}

\newcommand{\nickname}{PhysMani}

\newcommand{\myrightleftarrows}[1]{\mathrel{\substack{\xrightarrow{#1} \\[-.9ex] \xleftarrow{#1}}}}

\usepackage{xcolor}
\usepackage{color,soul}
\definecolor{bronze}{rgb}{1,1,0.6}
\definecolor{silver}{rgb}{0.969,0.796,0.600}
\definecolor{gold}{rgb}{0.941,0.592,0.600}
\newcommand{\gold}[1]{\colorbox{gold}{{#1}}}
\newcommand{\silver}[1]{\colorbox{silver}{{#1}}}

\begin{document}

\title{\nickname{}: Physics-principled 3D World Model for Dynamic Object Manipulation} 

\titlerunning{\nickname{}}

\author{Peng Yun\inst{1}\orcidlink{0000-0002-8163-267X} \and
Shouwang Huang\inst{1}\orcidlink{0009-0009-9655-2187} \and
Hao Li\inst{1}\orcidlink{0009-0003-4780-3767}
\and
Jinxi Li\inst{1}\orcidlink{0000-0001-9683-1359}
\and
Jianan Wang\inst{2}\orcidlink{0000-0003-2915-7960}
\and
Bo Yang\inst{1}\textsuperscript{\Letter}\orcidlink{0000-0002-2419-4140}
}

\authorrunning{P.~Yun et al.}

\institute{vLAR Group, The Hong Kong Polytechnic University, Hong Kong SAR, China
\and
Astribot, China  \quad \textsuperscript{\Letter} corresponding author
\\
\email{\{peng-aae.yun,bo.yang\}@polyu.edu.hk}}

\maketitle

\begin{abstract}
Manipulating fast and dynamically moving targets in unstructured 3D environments remains challenging for embodied AI.
Existing visual-language-action models and world models struggle with accurate 3D geometry and physically meaningful forecasting.
We propose \textbf{\nickname{}}, a framework that couples a physics-principled 3D Gaussian world model with a future-aware action policy model. The world model learns a divergence-free Gaussian velocity field via online optimization for fast and physically grounded future dynamics prediction. The policy model integrates the predicted 3D scene future dynamics through a learnable token based cross-attention module. We introduce \nickname{}-Bench, a dynamic manipulation benchmark with 16 tasks, and demonstrate a superior success rate over strong baselines in both simulation and real-world robot experiments.
Our code and data are available at \textcolor{red}{\href{https://github.com/vLAR-group/PhysMani}{https://github.com/vLAR-group/PhysMani}}
\end{abstract}

\section{Introduction}\label{sec:intro}
 
With the advancement of visual-language-action models (VLAs) \cite{Ma2025,Kawaharazuka2025} and world models \cite{Zhu2024d,Li2025,Ding2025}, embodied AI has seen tremendous progress in recent years, enabling robots to learn complex behaviors across diverse tasks. While achieving excellent performance, these models predominantly focus on static or quasi-static tasks, leaving the manipulation of dynamic objects underexplored. 

To date, only a few works have begun to address the challenge of manipulating dynamic objects, such as playing table tennis \cite{Dambrosio2025}, playing soccer \cite{Kitano1997}, or grasping moving objects \cite{Marturi2019,Noh2025} on conveyor belts \cite{Akinola2021,Zhou2024,Zhang2025} or table surfaces \cite{Xie2026}, \etc{} Despite achieving encouraging results, they are often limited to either specific or oversimplified scenarios. As a result, they fail to provide a general framework for dynamic object manipulation in open and unstructured environments, where objects or targets may exhibit complex motion patterns. 

In general dynamic interaction scenarios, such as catching a thrown ball, loading an item into a moving container, or placing a plate onto a rotating table, a robot must quickly and accurately anticipate future dynamics, which are governed by physical laws, and use those predictions to guide precise actions, all in 3D space. To accomplish such tasks, a potential strategy is to leverage existing powerful VLAs \cite{Sapkota2025} or video-based world models \cite{Ding2025} to jointly predict future states and action policies. However, these models currently fall short in three critical aspects: 1) their learned world representations often lack explicit awareness of 3D scene geometry; 2) their generated future frames, while visually realistic, frequently fail to capture physically meaningful dynamics \cite{Motamed2025}; 3) their chained large visual and language models often incur significant inference latency, which is prohibitive for time-sensitive dynamic manipulation.  

\begin{figure*}[t]
\centering
\includegraphics[width=1.0\linewidth]{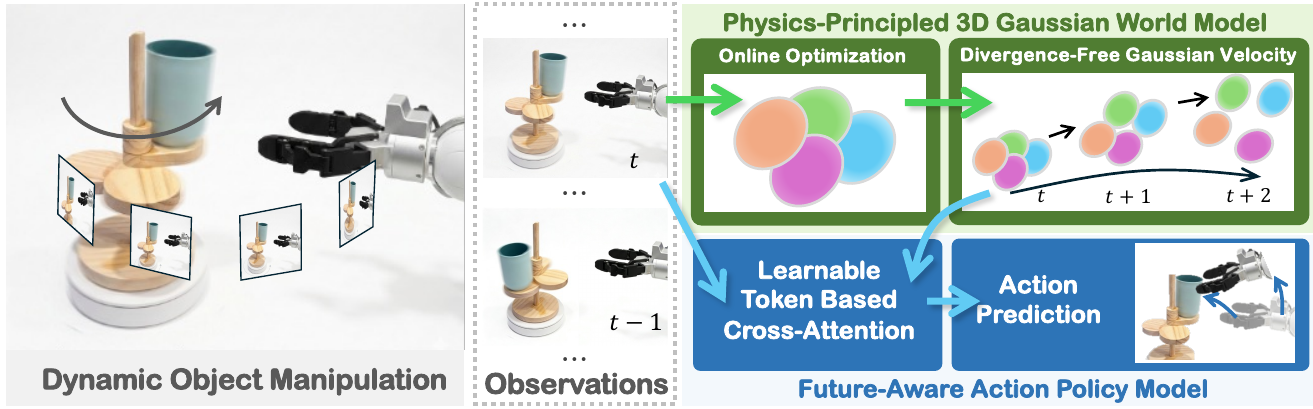}
\vskip -0.1in
\caption{The overall framework of our method. It features a physics-principled 3D Gaussian world model and a future-aware action policy model.}
\label{fig:overview}
\vskip -0.25in
\end{figure*}

To tackle these issues, we propose \textbf{\nickname{}}, a new framework for manipulating dynamic targets in general environments. As illustrated in Figure \ref{fig:overview}, our framework features two key designs: 1) a \textbf{physics-principled 3D Gaussian world model} that continuously optimizes and forecasts future 3D scene dynamics governed by fundamental physical laws; 2) a \textbf{future-aware action policy model} that seamlessly integrates predicted future dynamics into decision-making. These two components are realized in a lightweight pipeline that enables rapid future prediction and low-latency action execution.

For our physics-principled 3D world model, we build upon recent advances in 3D Gaussian-based physics learning, notably FreeGave \cite{li2025freegave}. Particularly, we introduce an online optimization mechanism that efficiently predicts a divergence-free per-Gaussian velocity field, thereby guaranteeing that the anticipated future scene dynamics adhere to fundamental physical laws.

For our future-aware action policy model, we adopt a simple yet effective cross-attention mechanism with learnable tokens to seamlessly integrate the predicted 3D scene future dynamics into decision-making. Specifically, for each 3D Gaussian in the dynamic environment, we attentively incorporate the estimated velocity information from its local spatial neighborhood into the policy network, so that the robot's predicted future actions are explicitly conditioned on the anticipated future motion of dynamic targets.
  
Since existing robotic benchmarks overwhelmingly focus on static or quasi-static manipulation and research on dynamic target manipulation is still in its infancy, we introduce \textbf{\nickname{}-Bench}, a moderate-sized dynamic manipulation benchmark with 16 tasks, inspired by the widely-used static benchmark RLBench \cite{James2020}, to evaluate our method. Overall, our contributions are:
\begin{itemize}[leftmargin=*] \vspace{-0.2cm}
\setlength{\itemsep}{1pt}
\setlength{\parsep}{1pt}
\setlength{\parskip}{1pt}
    \item We introduce a new framework and the first benchmark with 16 tasks for manipulating targets in general dynamic scenarios. 
    \item Our framework features a physics-principled 3D Gaussian world model for forecasting future dynamics and a future-aware action policy model that integrates predicted future dynamics.  
    \item We demonstrate superior manipulation success rates on both our newly introduced simulation benchmark and real-world experiments, clearly outperforming all baselines in terms of effectiveness and efficiency. \vspace{-0.2cm}
\end{itemize}

\section{Related Work}\label{sec:liter}

\phantom{XX}\textbf{VLAs and World Models}:
By leveraging advances in large language models (LLMs) \cite{OpenAI2023,Llama2024,DeepSeek-AI2024,Yang2025} and vision-language models (VLMs) \cite{Liu2023b,Qwen2025}, vision-language-action models (VLAs) \cite{Brohan2023,Ghosh2024,Kim2024b,Black2024,Zhang2025b,Kawaharazuka2025} have recently attracted significant attention in robotics. By jointly learning visual, language, and action modalities in an end-to-end framework, VLAs aim to enable robots to perform a wide range of tasks across diverse scenarios. Powered by Transformer architectures \cite{Vaswani2017} and/or diffusion models \cite{Ho2020}, VLAs have achieved strong success rates on many everyday static or quasi-static tasks. However, they often lack a comprehensive understanding of action and world dynamics, which limits their ability to perform foresight planning necessary for dynamic robotic tasks.

By learning to forecast future states conditioned on current observations and actions, world models \cite{Ha2018,Hafner2025} aim to capture environment dynamics and thereby inform decision-making. Recently, rapid progress in large-scale video generation \cite{Blattmann2023,Bar-Tal2024,NVIDIA2025,Kong2025,Wan2025} has revealed their potential as world models \cite{Assran2025,Ali2025,bruce2024genie,Tang2025,Chen2025,Bi2025,ye2026worldactionmodelszeroshot}, largely due to powerful diffusion Transformer architectures. Although these models can produce visually compelling future frames, the predicted dynamics are limited to 2D image space and often violate basic physical laws, making them unreliable for precisely guiding robotic actions in 3D space. Moreover, the use of chained large models typically results in slow inference, which hinders rapid forecasting and action execution in dynamic interaction scenarios.

\textbf{Manipulation of Dynamic Targets}: The recent VLAs and world models predominantly focus on static and quasi-static robotic tasks, leaving the manipulation of dynamic targets largely underexplored. Early efforts on dynamic robotic manipulation are mainly narrowed on specific tasks, including playing soccer \cite{Kitano1997}, playing table tennis \cite{Dambrosio2025}, object handovers \cite{Marturi2019,Liu2023,Christen2023,Noh2025}, grasping moving objects on conveyor belts \cite{Akinola2021,Zhou2024,Zhang2025} or table surfaces \cite{Xie2026}, \etc{} While achieving promising results, these methods do not provide a general framework for tackling dynamic manipulation in broader settings, where objects often exhibit complex and diverse physical dynamics. Moreover, most existing benchmarks and datasets \cite{James2020,Mees2022,Liu2023c} focus on static or quasi-static tasks, inherently hampering the study of dynamic manipulation. 

\textbf{Visual Physics Learning for Robotics}: 
Obtaining an accurate physical model of complex, dynamic 3D environments is crucial for embodied manipulation, especially in highly interactive settings. One line of work \cite{Li2019,Barcellona2025,Li2025b} relies on physics simulators \cite{coumans2021} to compute object-centric dynamics from known physical properties, but such approaches are typically restricted to specific object categories and lack generality across diverse objects. More recently, another line of research \cite{Li2023c,li2025freegave,Li2025d} has focused on learning 3D dynamics, particularly 3D velocity fields, directly from videos without requiring additional priors, offering a promising avenue for robots to predict future dynamics that adhere to fundamental physical laws.

\section{\nickname{}}\label{sec:meth}
\begin{figure*}[t]
\centering
\includegraphics[width=1.0\linewidth]{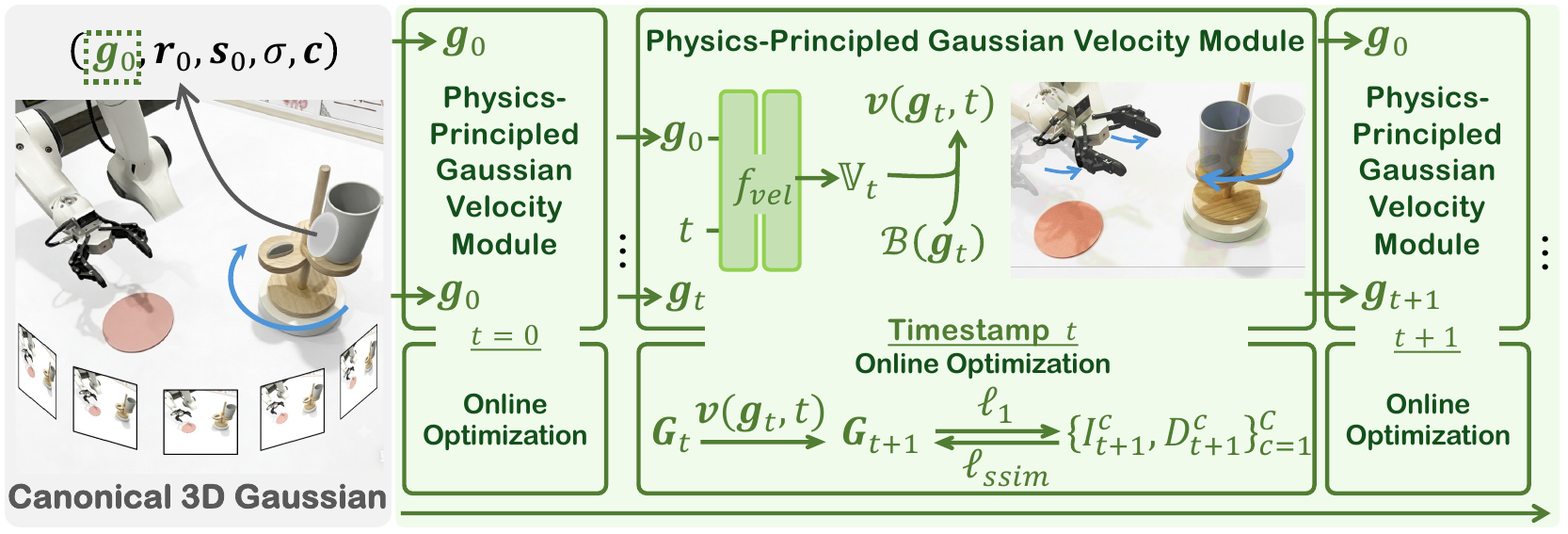}
\vskip -0.1in
\caption{The left panel shows the canonical 3D Gaussian module. The top-right panel shows the physics-principled Gaussian velocity module, and the bottom-right panel illustrates the online optimization process.}
\label{fig:world_model}
\vskip -0.15in
\end{figure*}

\subsection{Overview}

Our framework consists of two parallel models. The physics-principled 3D Gaussian world model is designed to continuously learn the 3D geometry, appearance, and, most importantly, physical dynamics of the target dynamic environment. For generality, we treat the embodiment as an intrinsic part of the dynamic environment rather than modeling it separately. While the world model predicts 3D scene future dynamics, our future-aware action policy model takes the current visual observations together with the predicted future dynamics as input and infers robot’s future actions for execution.

\subsection{Physics-principled 3D Gaussian World Model}\label{sec:phys_wm}

Given a dynamic environment for robot manipulation, our world model aims to learn both the 3D scene structure and dynamics. We draw inspiration from the recent FreeGave \cite{li2025freegave} whose key innovation is a divergence-free Gaussian velocity field for future frame prediction while respecting basic physical laws. However, its complex pipeline, relying on a separate deformation network and offline optimization, makes it impractical for real-time dynamic manipulation. To this end, we substantially refine FreeGave’s architecture and reformulate it as an online optimization model. For clarity and fluency, we present necessary details of the three main modules as follows. 

\textbf{Canonical 3D Gaussian Module}: As shown in the left block of Figure \ref{fig:world_model}, at time $t=0$, given RGB/Ds from $C$ cameras, denoted by $\{ (I^c_0, D_0^c) \}_{c=1}^C$, typically mounted at fixed and known poses in standard robot manipulation setups such as RLBench \cite{James2020} and CALVIN \cite{Mees2022}, we first initialize a set of 3D Gaussian kernels $\bm{G}_0$ to represent the canonical scene geometry and appearance at time $t=0$ based on the sparse point cloud back-projected from $C$ depth views. Each Gaussian is parameterized by a 3D position $\bm{g}_0$, scaling $\bm{s}_0$, covariance matrix converted from quaternion $\bm{r}_0$, opacity $\sigma$, and color $\bm{c}$. Following 3DGS\cite{Kerbl2023}, we then optimize these canonical kernels using the $C$ RGB images at time $t=0$ via $\ell_1$ and $\ell_{ssim}$ losses as used in 3DGS:   
\begin{equation}
\setlength{\abovedisplayskip}{3pt}
\setlength{\belowdisplayskip}{3pt}
    \underbrace{ \bigl\{\cdot\cdot (\bm{g}_0, \bm{r}_0, \bm{s}_0, \sigma, \bm{c}) \cdot\cdot \bigr\} }_{\bm{G}_0}  
    \mathrel{\mathop{\myrightleftarrows{\rule{1.cm}{0cm}}}^{\mathrm{render}}_{\mathrm{\ell_1 + \ell_{ssim}}}}
    \bigl\{I^1_0 \cdot\cdot I^c_0 \cdot\cdot I^C_0 \bigr\}
\end{equation}

\begin{algorithm}[t!]
\caption{ {\small Online Optimization of Future Dynamics.}
}
\label{alg:online_opt}
\begin{algorithmic} 
\footnotesize
\State{\textbf{Preliminary:}} Canonical 3D Gaussians: $\bigl\{\cdot\cdot (\bm{g}_0, \bm{r}_0, \bm{s}_0, \sigma, \bm{c}, \bm{z}) \cdot\cdot \bigr\}$; 

\While {true}
\State{$\bullet$ receiving new RGB/Ds, denoted by $\{I_t^c, D_t^c\}_{c=1}^C$, at time $t$, where $t \leftarrow t'+1 $; \\ 
\textit{\phantom{xx} - Note that, here $t'$ denotes the previous time step, while $t$ denotes the current time step, which may correspond to the immediately subsequent frame or to a frame after several intervals following $t'$.}
}

\State{$\bullet$ getting $\bm{v}(\bm{g}_t, t)$ from the Gaussian velocity module;
}

\State{$\bullet$ applying $\bm{v}(\bm{g}_t, t)$ to drive the corresponding Gaussian from $t'$ to $t$ using the same interleaved mid-point method proposed in FreeGave \cite{li2025freegave};
}

\State{$\bullet$ optimizing the Gaussian's position $\bm{g}_t$, orientation $\bm{r}_t$, and $f_{vel}$, based on $\ell_1$ and $\ell_{ssim}$ losses between rendered RGB/Ds and $\{I_t^c, D_t^c\}_{c=1}^C$. This gradient-based optimization lasts $T$ iterations considering the trade-off between accuracy and efficiency; \\

\textit{\phantom{xx} - Note that the above optimization stage is primarily conducted for learning physically meaningful basic components $\mathbb{V}_t$ and velocity field $\bm{v}(\bm{g}_t, t)$ for each Gaussian.}
}

\State{$\bullet$ refining the Gaussian properties for a few additional $T'$ iterations using the same $\ell_1$ and $\ell_{ssim}$ losses, while freezing the velocity network $f_{vel}$. This compensates for potential minor changes in scene geometry and appearance over time.  
}

\State{$\bullet$ \textbf{Output}: the six basic velocity components $\mathbb{V}_t$ and $\bm{v}(\bm{g}_t, t)$ for every Gaussian} 

\EndWhile

\State{\textit{\phantom{xx} - In our experiments, we set $T=50$ and $T'=7$ iterations. Each round of optimization takes $\sim 200ms$ on a single RTX4090 GPU.}}
\end{algorithmic}
\end{algorithm}

\textbf{Physics-principled Gaussian Velocity Module}: As shown in the top-right block of Figure \ref{fig:world_model}, for each Gaussian at current time $t$, we follow FreeGave to use MLPs, denoted by $f_{vel}$, to learn six basic velocity components: $\mathbb{V}_t\leftarrow[v_t^x, v_t^y, v_t^z, w_t^z, w_t^y, w_t^x]$, where $v_t^x / v_t^y / v_t^z$ represent linear velocities and $w_t^z / w_t^y / w_t^x$ angular velocities. The velocity field, denoted by $\boldsymbol{v}(\boldsymbol{g}_t, t)$, is composed as follows: \vspace{-0.3cm}
\begin{equation}\label{eq:vel_decomp}
\setlength{\abovedisplayskip}{-3pt}
\setlength{\belowdisplayskip}{3pt}
 \boldsymbol{v}(\boldsymbol{g}_t, t) = \mathbb{V}_t \cdot \mathcal{B}(\boldsymbol{g}_t), \quad \mathbb{V}_t = f_{vel}(\bm{g}_0,t), \quad \mathbb{V}_t \in \mathcal{R}^{1\times 6}\quad \mathcal{B}(\boldsymbol{g}_t) \in \mathcal{R}^{6\times 3}
 \vspace{-0.2cm}
\end{equation}
where $\mathcal{B}(\bm{g}_t)$ is a set of basis vectors and $g_t^x / g_t^y / g_t^z$ are coordinates of $\bm{g}_t$: 
\begin{equation}\label{eq:divfree_basis}
\setlength{\abovedisplayskip}{3pt}
\setlength{\belowdisplayskip}{3pt}
\mathcal{B}(\bm{g}_t) = 
\begin{bmatrix}
1  &0  &0  &-g_t^y  &g_t^z  &0 \\
0  &1  &0  &g_t^x  &0  &-g_t^z \\
0  &0  &1  &0  &-g_t^x &g_t^y \\
\end{bmatrix}^T
\end{equation}\vspace{-0.15cm}

A key advantage of this Gaussian velocity module is that it enforces a divergence-free property, as proved in the FreeGave paper.

\textbf{Online Optimization of Future Dynamics}: In the context of dynamic manipulation, a stream of RGB/Ds is continuously available, and our world model must be optimized rapidly and online so that it can capture the ongoing dynamics of the target and environment, and accurately forecast future evolution. To achieve this, we remove the auxiliary deformation field and the deformation-aided optimization strategy in FreeGave \cite{li2025freegave}. Instead, we propose the following online optimization Algorithm \ref{alg:online_opt} to learn precise velocity components $\mathbb{V}_t$ and field $\bm{v}(\bm{g}_t, t)$ for every Gaussian at time $t$. More details about the network and optimization of our world model are provided in Appendix A.1.

\subsection{Future-aware Action Policy Model}\label{sec:action_policy_model}
Our action policy model aims to leverage both the current visual observations and the predicted future dynamics to infer future actions. For generality, we adopt the existing 3D FlowMatch Actor (3DFA) \cite{Gkanatsios2025} as our policy network backbone, though others could also be used. In fact, 3DFA is built on 3D Diffuser Actor (3DDA) \cite{Ke2024} by replacing DDPM-based diffusion with Rectified Flow \cite{Bordes2023} for fast training and inference, without touching the network architecture. Our policy model consists of the following components.

\begin{figure*}[t]
\centering
\includegraphics[width=1.0\linewidth]{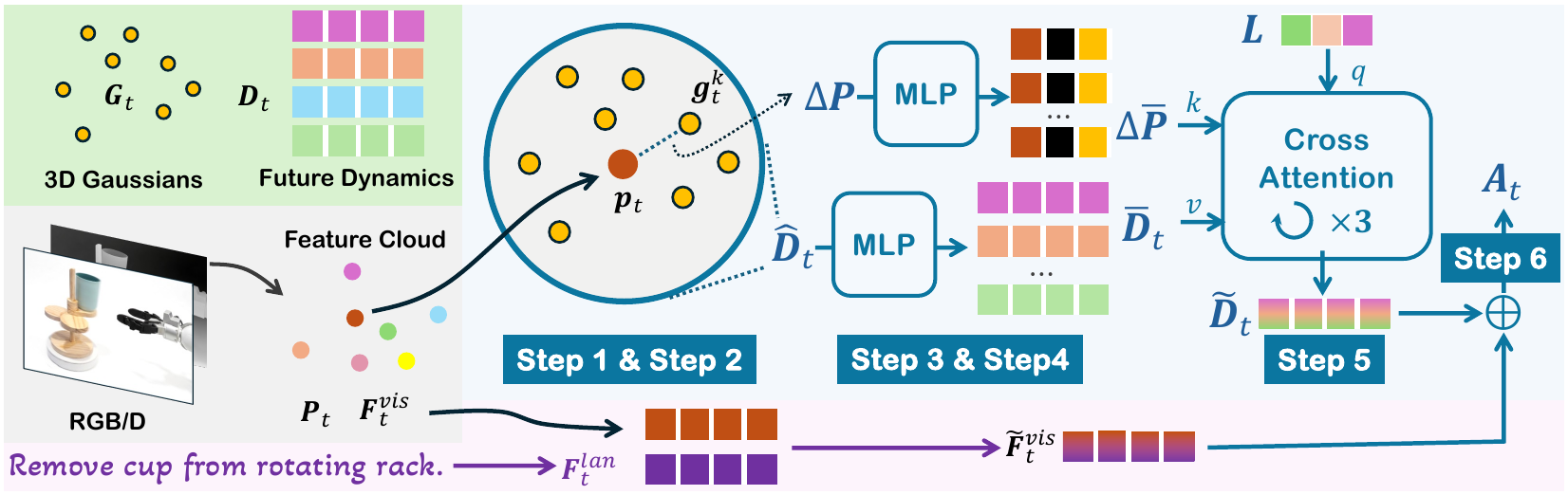}
\vskip -0.1in
\caption{The left block shows the encoding of visual observations and language. The right block shows the incorporation of 3D scene future dynamics for action prediction.}
\label{fig:policy_model}
\vskip -0.15in
\end{figure*}

\textbf{Encoding 3D Visual Geometry, Appearance and Language}: As illustrated in the left block of Figure \ref{fig:policy_model}, we follow 3DFA \cite{Gkanatsios2025} to process the posed RGB/Ds at current time $t$, lifting 2D visual features to a sparse 3D scene point cloud $\bm{P}_t\in\mathcal{R}^{4096\times 3}$ converted from depth views, getting a feature cloud or a cloud of visual tokens, denoted by $\bm{F}_t^{vis}\in \mathcal{R}^{4096\times 120}$. We also encode the language task instruction into a set of tokens, denoted by $\bm{F}_t^{lan}\in\mathcal{R}^{53\times120}$. Eventually, these two sets of tokens are fused via cross-attention, obtaining a cloud of new tokens, denoted by $\bm{\tilde{F}}_t^{vis}\in\mathcal{R}^{4096\times120}$. More details can be found in 3DFA.

\textbf{Incorporating 3D Scene Future Dynamics}: At current time $t$, our world model predicts six basic velocity components $\mathbb{V}_t$ for each Gaussian, yielding the future dynamics of the entire 3D scene, denoted by $\mathbf{D}_t\in\mathcal{R}^{H\times6}$, where $H$ is the total number of 3D scene Gaussians $\bm{G}_t$. We set $H=30000$ in all experiments. 

As shown in the right block of Figure \ref{fig:policy_model}, we incorporate the scene dynamics $\mathbf{D}_t$ into visual tokens $\mathbf{\tilde{F}}_t^{vis}$ of the sparse 3D scene point cloud $\bm{P}_t$ as follows:
\begin{itemize}[leftmargin=*] \vspace{-0.2cm}
\setlength{\itemsep}{1pt}
\setlength{\parsep}{1pt}
\setlength{\parskip}{1pt}
    \item Step \#1: For each point $\bm{p}_t$ in the 3D scene point cloud $\bm{P}_t$, we retrieve its $K$ nearest neighboring 3D Gaussians from our world model via KNN, based on Euclidean distance between point and Gaussian coordinates.  
    \item Step \#2: For each of the $K$ nearest Gaussians $\{\bm{g}_t^1 \cdots \bm{g}_t^k \cdots \bm{g}_t^K\}$ associated with the center point $\bm{p}_t$, we compute the relative offset: $\Delta \bm{p} = \bm{p}_t - \bm{g}_t^k$.
    \item Step \#3: For the entire 3D scene point cloud $\bm{P}_t$, we similarly compute relative offsets to all $K$ neighboring Gaussians, yielding $\Delta 
\bm{P} \in \mathcal{R}^{4096\times K \times 3}$, 
    and retrieve their basic velocity components, denoted by $\bm{\hat{D}}_t \in \mathcal{R}^{4096\times K \times 6}$.   
    \item Step \#4: Both $\Delta \bm{P}$ and $\bm{\hat{D}}_t$ are fed into two separate MLPs to obtain two sets of tokens, denoted by $\Delta \bm{\bar{P}} \in \mathcal{R}^{4096\times K \times 120}$ and $\bm{\bar{D}}_t \in \mathcal{R}^{4096\times K \times 120}$ respectively. 
    \item Step \#5: We introduce a learnable vector $\bm{L}\in\mathcal{R}^{1\times 120}$ to act as the query token, and regard $\Delta \bm{\bar{P}}$ and $\bm{\bar{D}}_t$ as key and value tokens respectively. These key, query, value tokens are fed into three attention blocks, obtaining the final value tokens $\bm{{\tilde{D}}}_t\in \mathcal{R}^{4096\times120}$. Basically, $\bm{{\tilde{D}}}_t$ serves as the future dynamics tokens corresponding to the 3D scene point cloud $\bm{P}_t$. 
    \item Step \#6: Lastly, we add dynamics tokens $\bm{{\tilde{D}}}_t$ to visual tokens $\mathbf{\tilde{F}}_t^{vis}$, obtaining future-aware tokens $\bm{A}_t\in\mathcal{R}^{4096\times120}$, which will be used to infer robot actions. \vspace{-0.8cm}
\end{itemize}
More details of these incorporation steps are provided in Appendix A.2. 

\textbf{Denoising End-effector's Keyposes with Rectified Flow}: Following the neural architecture of 3DFA \cite{Gkanatsios2025}, our action model also aims to predict the end-effector's keyposes, which is commonly adopted in prior works \cite{Shridhar2022,Goyal2023,James2022,Liu2022}. Particularly, each action keypose $\bm{a}$ consists of 3D position, 3D orientation, and a binary open/closed state of the end-effector: $\bm{a} = \{\bm{a}^{pos}\in\mathcal{R}^3, \bm{a}^{rot}\in\mathcal{R}^6, \bm{a}^{open}\in\{0,1\}\}$. At current timestep $t$, our goal is to predict the corresponding end-effector trajectory $\bm{\tau}_t = (\bm{a}^{pos}_{t:t+J}, \bm{a}^{rot}_{t:t+J})$ of temporal horizon $J$. 

Given a clean training trajectory, denoted by $\bm{\tau}^0$ where time step $t$ is ignored for simplicity, at a denoising step $i$, we create the noisy trajectory by a linear interpolation: $\bm{\tau}^i = (1-i)\bm{\epsilon} + i\bm{\tau}^0$, where $\bm{\epsilon}$ is the sampled noise. 

Our model takes as input the noisy trajectory $\bm{\tau}^i$, proprioception information $\bm{o}$, language task instruction $\bm{l}$, and most importantly, our future-aware 3D scene tokens $\bm{A}$ (subscript time $t$ is also ignored for simplicity), and outputs the velocity field $V_{flow}$ for flow matching and gripper openness $f_{open}$, respectively.  

During training, we follow 3DFA \cite{Gkanatsios2025} to sample the action time step $t$ uniformly in training trajectories, denoising time step $i \sim \sigma(\mathcal{N}(0,1))$, and noise $\bm{\epsilon} \sim \mathcal{N}(\bm{0}, \bm{1})$. We use the $\ell_2$ loss to supervise the velocity field for flow matching and binary-cross-entropy (BCE) to supervise the openness:
\begin{equation}
\setlength{\abovedisplayskip}{3pt}
\setlength{\belowdisplayskip}{3pt}
    \ell = \ell_2\Big(V_{flow}(\bm{A}, \bm{o}, \bm{l}, \bm{\tau}^i, i), (\bm{\epsilon} - \bm{\tau}^0) \Big) + BCE \Big( f_{open}(\bm{A}, \bm{o}, \bm{l}, \bm{\tau}^i, i), \bm{a}^{open}_{t:t+J} \Big) 
\end{equation}

During inference, given visual observations, our framework continuously predicts the 3D scene’s future dynamics and uses them to infer future action keyposes, which are then converted into joint commands via inverse kinematics for execution. More details of the network and training are in Appendix A.2.

\subsection{Implementation for Fast Computation}
To speed up the overall computation of our framework, we reuse static computation graphs during online optimization of future dynamics. In our setting, the number of 3D Gaussians remains fixed across frames,  allowing the optimization to be executed within a static computation graph.
By avoiding dynamic graph construction, we significantly reduce CUDA kernel launch overhead and improve GPU utilization. More details of the kernel launch time and overall update latency with and without reusing static graph is in Appendix A.3.

\section{\nickname{}-Bench}\label{sec:bench}
\begin{figure*}[t]
\centering
\includegraphics[width=.99\linewidth]{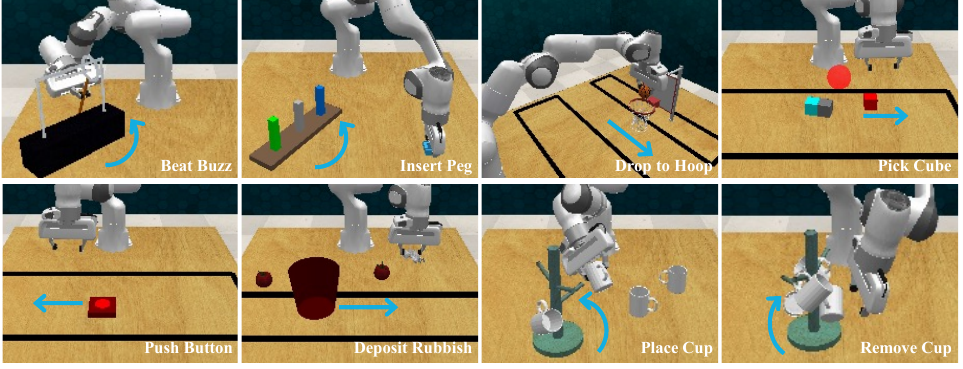}
\vskip -0.1in
\caption{Examples of diverse and challenging dynamic tasks in our \nickname{}-Bench. The blue arrows illustrate the complex motions of dynamic targets.}
\label{fig:bench_example}
\vskip -0.15in
\end{figure*}

As there is a lack of benchmarks for manipulating targets in general dynamic settings, we carefully extend the widely used RLBench \cite{James2020} and construct 8 groups of general dynamic tasks, each with normal-speed and high-speed variants, resulting in 16 challenging tasks in total, as briefly described below and illustrated in Figure \ref{fig:bench_example}. These tasks cover a diverse range of dynamic reaching, grasping, placing, inserting, pushing and tool-use, with dynamic targets exhibiting complex motions such as linear, circular, and rotational trajectories. While evaluating on a very large-scale benchmark with hundreds or even thousands of tasks is desirable, it is non-trivial. Although recent PhysInOne \cite{Zhou2026} has begun to realize this scale for visual physics reasoning, scaling our physics-principled evaluation framework to a comparable magnitude is left for future work.

\begin{itemize}[leftmargin=*] \vspace{-0.2cm}
\setlength{\itemsep}{1pt}
\setlength{\parsep}{1pt}
\setlength{\parskip}{1pt}
    \item \textit{beat the rotating buzz at normal/high speed}: The robot must grasp and steer the loop along the rotating wire, maintaining position without making contact.
    \item \textit{insert ring onto rotating peg at normal/high speed}: The robot must pick up a ring, align and insert it onto a vertical peg attached to a rotating base. 
    \item \textit{drop basketball into moving hoop at normal/high speed}: The robot must pick up a ball and then drop it into a hoop attached to a moving backboard. 
    \item \textit{pick moving cube on table at normal/high speed}: The robot must pick up a moving cube among several other static cubes on the table.
    \item \textit{push moving button at normal/high speed}: The robot must press a button moving on the table.
    \item \textit{deposit rubbish into moving bin at normal/high speed}: The robot must pick up the rubbish among other items and deposit it into a moving bin.
    \item \textit{place cup onto rotating rack at normal/high speed}: The robot must pick up a cup and hang it on the rotating rack. 
    \item \textit{remove cup from rotating rack at normal/high speed}: The robot must grasp a cup from the rotating rack and place it on the table.
    \vspace{-0.2cm}
\end{itemize}

For every task, 4 static cameras are mounted around the table, the same as RLBench. The speeds of moving targets are set comparable to the robot’s maximum end-effector speed (up to 2 m/s for the Franka Emika Panda 7‑DoF arm). This dynamic scenario is highly challenging; reactive control alone is inadequate.

For each task, we collect 160 successful trajectories, randomly split them into training/validation/test sets in a 100/20/40 ratio. The trajectories are gathered using scripted expert policies which produce collision-free and valid executions. All methods are trained on the same training split and evaluated online in simulation using identical configurations in the test split. More details of our \nickname{}-Bench are in Appendix A.4.  

\section{Experiments}\label{sec:exp}

We compare against the following representative baselines and evaluate the success rate (\textbf{SR}) and future frame prediction quality as primary metrics:
\begin{itemize}[leftmargin=*] \vspace{-0.1cm}
\setlength{\itemsep}{1pt}
\setlength{\parsep}{1pt}
\setlength{\parskip}{1pt}
    \item \textbf{Act3D} \cite{Gervet2023}: It is a voxel-based 3D action prediction method that formulates manipulation as a sequence of perception-driven affordance predictions.
    \item \textbf{3DDA} \cite{Ke2024}: It is a diffusion-based imitation learning policy that generates continuous action trajectories conditioned on 3D scene representations. For a fair comparison, we adopt 10 denoising steps to predict its policies, matching the same time budget used by 3DFA and our method. 
    \item \textbf{3DFA} \cite{Gkanatsios2025}: It improves 3DDA by replacing DDPM-based diffusion with Rectified Flow for fast training and inference.
    \item \textbf{3DFA-OF}: A motion-cue variant of 3DFA in which 2D optical flow is concatenated with the RGB input (5 channels in total) and projected back to 3 channels by a convolutional layer before the 3DFA encoder.
    \item \textbf{ManiGaussian} \cite{lu2024manigaussian}: It is based on dynamic Gaussian splatting to learn 3D scene dynamics as an auxiliary task for improving action prediction.
    \item \textbf{pi0.5} \cite{Black2025}: It is a compact and sophisticated VLA model that predicts action chunks from visual observations and task instructions. \vspace{-0.2cm}
\end{itemize}

Since our goal is to manipulate targets across general and diverse dynamic scenarios, we evaluate all 16 tasks using a single model of each method rather than training separate models for each task. Specifically, we fine-tune pi0.5 on the entire training set with LoRA \cite{Hu2022}. All baselines and our method are well-trained using the same training set of \nickname{}-Bench in the same action space, ensuring a fair comparison. For each method, we report success rate (SR) and the standard deviation using the best 3 models of its last five checkpoints in Tables \ref{tab:res_bench}\&\ref{tab:res_future_frame}. 
More implementation details can be found in the Appendix A.2.
\subsection{Evaluation on \nickname{}-Bench}\label{sec:exp_bench}

\begin{table*}[t]\tabcolsep= 0.1cm 
    \centering
    \caption{The average success rate (SR) (\%) across all 16 tasks of our \nickname{}-Bench. The standard deviation is computed over the top 3 models of the last 5 checkpoints.}\vspace{-0.3cm}
    \label{tab:res_bench}
    \resizebox{\textwidth}{!}{
    \begin{tabular}{lccccccccc}
    \toprule 
    & \multirow{2}{*}{\begin{tabular}[c]{@{}c@{}} \textbf{Mean} \\ \textbf{SR}\end{tabular}}
    & \multirow{2}{*}{\begin{tabular}[c]{@{}c@{}} \textbf{Beat} \\ \textbf{Buzz}\end{tabular}} 
    & \multirow{2}{*}{\begin{tabular}[c]{@{}c@{}} \textbf{Insert} \\ \textbf{Peg}\end{tabular}} 
    & \multirow{2}{*}{\begin{tabular}[c]{@{}c@{}} \textbf{Drop to} \\ \textbf{Hoop}\end{tabular}} 
    & \multirow{2}{*}{\begin{tabular}[c]{@{}c@{}} \textbf{Pick} \\ \textbf{Cube}\end{tabular}} 
    & \multirow{2}{*}{\begin{tabular}[c]{@{}c@{}} \textbf{Push} \\ \textbf{Button}\end{tabular}}
    & \multirow{2}{*}{\begin{tabular}[c]{@{}c@{}} \textbf{Deposit} \\ \textbf{Rubbish}\end{tabular}}
    & \multirow{2}{*}{\begin{tabular}[c]{@{}c@{}} \textbf{Place} \\ \textbf{Cup}\end{tabular}} 
    & \multirow{2}{*}{\begin{tabular}[c]{@{}c@{}} \textbf{Remove} \\ \textbf{Cup}\end{tabular}} \\
    & & & & & & & & & \\
    \midrule
    Act3D \cite{Gervet2023} & $27.1 \pm 2.7$ & $46.7 \pm 5.9$ & \silver{$58.3 \pm 4.7$} & $6.7 \pm 1.2$ & $56.7 \pm 6.6$ & $49.2 \pm 11.2$ & \gold{$81.7 \pm 5.1$} & $0.0 \pm 0.0$ & $21.7 \pm 8.5$ \\
    pi0.5 \cite{Black2025} & $8.3 \pm 0.2$ & $16.7 \pm 3.1$ & $0.8 \pm 1.2$ & $0.8 \pm 1.2$ & $18.3 \pm 10.5$ & $34.2 \pm 4.7$ & $8.3 \pm 1.2$ & $0.0 \pm 0.0$ & $22.5 \pm 6.1$ \\
    ManiGaussian \cite{lu2024manigaussian} & $22.5 \pm 0.8$ & $40.0 \pm 8.9$ & $35.8 \pm 4.2$ & $6.7 \pm 6.2$ & $32.5 \pm 12.7$ & $18.3 \pm 2.4$ & \silver{$67.5 \pm 2.0$} & $1.7 \pm 1.2$ & $61.7 \pm 10.5$ \\
    3DDA \cite{Ke2024} & $35.1 \pm 1.7$ & $60.0 \pm 4.1$ & \gold{$68.3 \pm 6.6$} & $9.2 \pm 2.4$ & $63.3 \pm 6.6$ & $39.2 \pm 6.6$ & $67.5 \pm 4.1$ & \silver{$15.8 \pm 3.1$} & $58.3 \pm 3.1$ \\
    3DFA \cite{Gkanatsios2025} & \silver{$37.8 \pm 0.9$} & \gold{$82.5 \pm 2.0$} & $38.3 \pm 6.2$ & \silver{$30.8 \pm 3.1$} & \silver{$70.0 \pm 6.1$} & \silver{$55.8 \pm 4.2$} & $53.3 \pm 4.2$ & \gold{$17.5 \pm 4.1$} & \gold{$63.3 \pm 2.4$} \\
    3DFA-OF & $37.5 \pm 1.0$ & \silver{$79.2 \pm 4.7$} & $45.0 \pm 5.4$ & $27.5 \pm 5.4$ & $60.8 \pm 5.9$ & $54.2 \pm 3.1$ & $58.3 \pm 4.2$ & $14.2 \pm 5.1$ & $58.3 \pm 11.2$ \\
    \textbf{\nickname{}(Ours)} & \gold{$45.9 \pm 0.8$} & $78.3 \pm 4.2$ & $37.5 \pm 6.1$ & \gold{$71.7 \pm 3.1$} & \gold{$84.2 \pm 3.1$} & \gold{$57.5 \pm 3.5$} & $60.0 \pm 2.0$ & $15.0 \pm 3.5$ & \silver{$61.7 \pm 6.6$} \\
    \midrule
     
    & \multirow{3}{*}{\begin{tabular}[c]{@{}c@{}} \textit{High} \\ \textit{Speed} \\ \textbf{(H)$\xrightarrow{}$} \end{tabular}}
    & \multirow{3}{*}{\begin{tabular}[c]{@{}c@{}} \textbf{Beat} \\ \textbf{Buzz} \\ \textbf{(H)} \end{tabular}} 
    & \multirow{3}{*}{\begin{tabular}[c]{@{}c@{}} \textbf{Insert} \\ \textbf{Peg} \\ \textbf{(H)} \end{tabular}} 
    & \multirow{3}{*}{\begin{tabular}[c]{@{}c@{}} \textbf{Drop to} \\ \textbf{Hoop} \\ \textbf{(H)} \end{tabular}} 
    & \multirow{3}{*}{\begin{tabular}[c]{@{}c@{}} \textbf{Pick} \\ \textbf{Cube} \\ \textbf{(H)} \end{tabular}} 
    & \multirow{3}{*}{\begin{tabular}[c]{@{}c@{}} \textbf{Push} \\ \textbf{Button} \\ \textbf{(H)} \end{tabular}}
    & \multirow{3}{*}{\begin{tabular}[c]{@{}c@{}} \textbf{Deposit} \\ \textbf{Rubbish} \\ \textbf{(H)} \end{tabular}}
    & \multirow{3}{*}{\begin{tabular}[c]{@{}c@{}} \textbf{Place} \\ \textbf{Cup} \\ \textbf{(H)} \end{tabular}} 
    & \multirow{3}{*}{\begin{tabular}[c]{@{}c@{}} \textbf{Remove} \\ \textbf{Cup} \\ \textbf{(H)} \end{tabular}} \\
    & & & & & & & & & \\
    & & & & & & & & & \\
    \midrule
    Act3D \cite{Gervet2023} & -- & $19.2 \pm 5.9$ & $8.3 \pm 3.1$ & $3.3 \pm 1.2$ & $12.5 \pm 2.0$ & \gold{$25.8 \pm 11.6$} & $31.7 \pm 16.6$ & $0.0 \pm 0.0$ & $12.5 \pm 2.0$ \\
    pi0.5 \cite{Black2025} & -- & $3.3 \pm 1.2$ & $0.0 \pm 0.0$ & $0.0 \pm 0.0$ & $5.8 \pm 1.2$ & $19.2 \pm 2.4$ & $0.8 \pm 1.2$ & $0.0 \pm 0.0$ & $1.7 \pm 2.4$ \\
    ManiGaussian \cite{lu2024manigaussian} & -- & $22.5 \pm 3.5$ & $1.7 \pm 1.2$ & $0.8 \pm 1.2$ & $6.7 \pm 4.7$ & $1.7 \pm 1.2$ & \silver{$37.5 \pm 5.4$} & $0.0 \pm 0.0$ & $25.0 \pm 5.4$ \\
    3DDA \cite{Ke2024} & -- & $45.8 \pm 4.2$ & \silver{$25.0 \pm 10.2$} & $2.5 \pm 2.0$ & $10.0 \pm 2.0$ & $20.0 \pm 7.4$ & $25.8 \pm 5.9$ & $15.0 \pm 0.0$ & $36.7 \pm 7.2$ \\
    3DFA \cite{Gkanatsios2025} & -- & \gold{$52.5 \pm 9.4$} & $11.7 \pm 5.1$ & \silver{$9.2 \pm 3.1$} & \gold{$21.7 \pm 2.4$} & \silver{$21.7 \pm 3.1$} & $19.2 \pm 4.2$ & $13.3 \pm 3.1$ & $43.3 \pm 6.2$ \\
    3DFA-OF & -- & $48.3 \pm 8.5$ & $20.8 \pm 4.7$ & $6.7 \pm 2.4$ & \silver{$20.8 \pm 3.1$} & $15.8 \pm 6.2$ & $28.3 \pm 3.1$ & \silver{$18.3 \pm 6.2$} & \gold{$44.2 \pm 5.9$} \\
    \textbf{\nickname{}(Ours)} & -- & \silver{$49.2 \pm 7.7$} & \gold{$25.0 \pm 2.0$} & \gold{$46.7 \pm 4.2$} & $18.3 \pm 2.4$ & $20.0 \pm 5.4$ & \gold{$46.7 \pm 4.2$} & \gold{$19.2 \pm 4.2$} & \silver{$43.3 \pm 3.1$} \\
    \bottomrule
    \end{tabular}
    }\vspace{-0.35cm}
\end{table*}

\phantom{xW}\textbf{Dynamic Manipulation:}
In Table \ref{tab:res_bench}, \nickname{} achieves the best performance across most tasks, obtaining a mean SR score 8.1 points higher than the next best method 3DFA. Notably, \nickname{} uses 3DFA as its backbone and adopts the same training strategy, differing only in its incorporation of 3D scene future dynamics. It shows the effectiveness of \nickname{} with the performance gains directly attributed to our accurate future dynamics predictions (Table \ref{tab:res_future_frame}).

Augmenting 3DFA with 2D optical flow as an explicit motion cue (3DFA-OF) yields no performance gain over the baseline 3DFA and still falls significantly short of \nickname{}. This indicates that 2D motion cues alone provide limited dynamic information. Instead, the substantial improvements seen in \nickname{} stem from integrating the future dynamics generated by our world model.

The strong VLA model pi0.5 performs poorly on almost all high-speed variants, mainly because it fails to capture the 3D spatial representation and the complex future dynamics required for precise manipulation.

\textbf{Future Frame Prediction:}
Table~\ref{tab:res_future_frame} reports PSNR, SSIM, LPIPS, logRMSE (on depth images) and trajectory error to evaluate future frame prediction for our method and ManiGaussian; the other baselines in Table~\ref{tab:res_bench} do not support this. 
We further compare our method with FreeGave \cite{li2025freegave}, evaluating it online using current and historical observations within the same iteration budget.

\begin{figure}[t]
    \centering
    \includegraphics[width=1\linewidth]{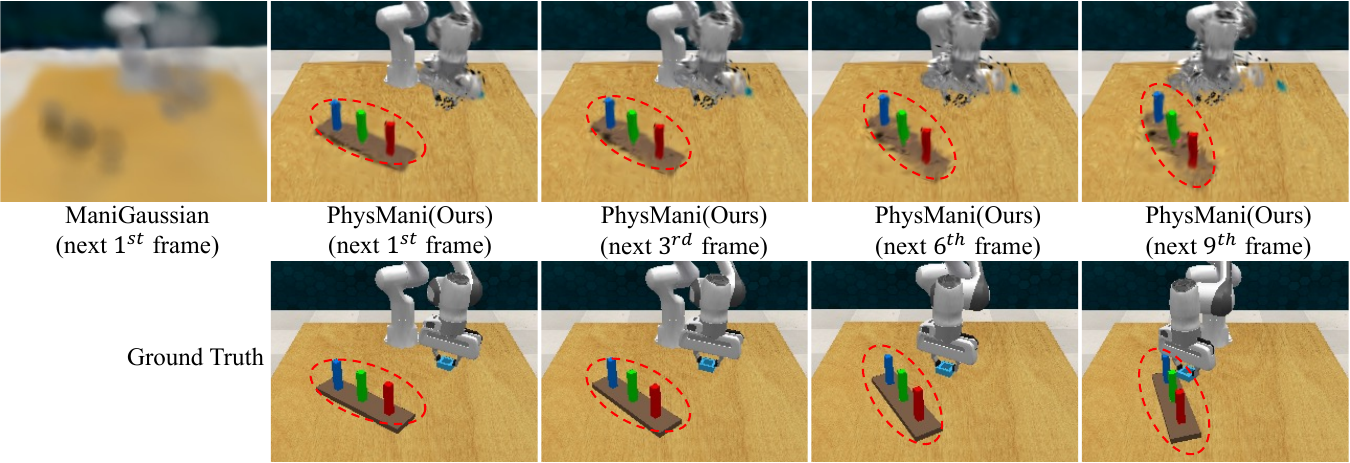}
    \caption{Qualitative comparison of future frame prediction. Red circles highlight that our method can accurately predict the movement of dynamic targets.}
    \label{fig:res_future_frame_qualitative}
    \vspace{-0.5cm}
\end{figure}

For every timestep of each test trajectory across 16 dynamic tasks, \nickname{} renders RGB/D predictions from all four cameras for the next 10 timesteps. Since ManiGaussian predicts only one future frame, we render only the next timestep. All predictions are compared
with ground truth. Trajectory error assesses whether predictions capture physically meaningful motion, \ie, the Euclidean distance between predicted and ground-truth target centers.
Although ManiGaussian is designed for future-frame prediction, it performs relatively poorly because its predictions are inaccurate (PSNR 14.32; Table~\ref{tab:res_future_frame} and Figure~\ref{fig:res_future_frame_qualitative}).  In contrast, our physics-principled module enables high-quality future-frame predictions over 10 timesteps.

\begin{table*}[t]\tabcolsep= 0.55cm
    \centering
    \caption{Quantitative comparison of future frame prediction on \nickname{}-Bench.}\vspace{-0.3cm}
    \label{tab:res_future_frame}
    \resizebox{\textwidth}{!}{
    \begin{tabular}{lccccc}
        \toprule
        & \textbf{PSNR} $\uparrow$ & \textbf{logRMSE} $\downarrow$ & \textbf{SSIM} $\uparrow$ & \textbf{LPIPS} $\downarrow$ & \textbf{Traj. Err.} $\downarrow$ \\
        \midrule
            ManiGaussian \cite{lu2024manigaussian} (next $1^{st}$ frame) & 14.32 & - & 0.402 & 0.889 & 0.388 \\
            FreeGave \cite{li2025freegave} (next $1^{st}$ frame) & 19.47 & 0.208 & 0.799 & 0.208 & 0.010 \\
            \textbf{\nickname{}(Ours)} (next $1^{st}$ frame) & \gold{$26.90$} & \gold{$0.060$} & \gold{$0.877$} & \gold{$0.123$} & \gold{$0.008$} \\
        \midrule
            FreeGave \cite{li2025freegave} (next $5^{th}$ frame) & 18.33 & 0.253 & 0.780 & 0.234 & 0.047 \\
            \textbf{\nickname{}(Ours)} (next $5^{th}$ frame) & \gold{$22.42$} & \gold{$0.134$} & \gold{$0.829$} & \gold{$0.157$} & \gold{$0.039$} \\
        \midrule
            FreeGave \cite{li2025freegave} (next $10^{th}$ frame) & 17.89 & 0.269 & 0.770 & 0.250 & 0.086 \\
            \textbf{\nickname{}(Ours)} (next $10^{th}$ frame) & \gold{$20.00$} & \gold{$0.189$} & \gold{$0.786$} & \gold{$0.196$} & \gold{$0.074$} \\
        \bottomrule
    \end{tabular}}
\vspace{0.35cm}
\end{table*}

\begin{figure}[h]
    \vspace{-0.5cm}
    \centering
    \setbox0\hbox{\includegraphics[width=0.66\textwidth]{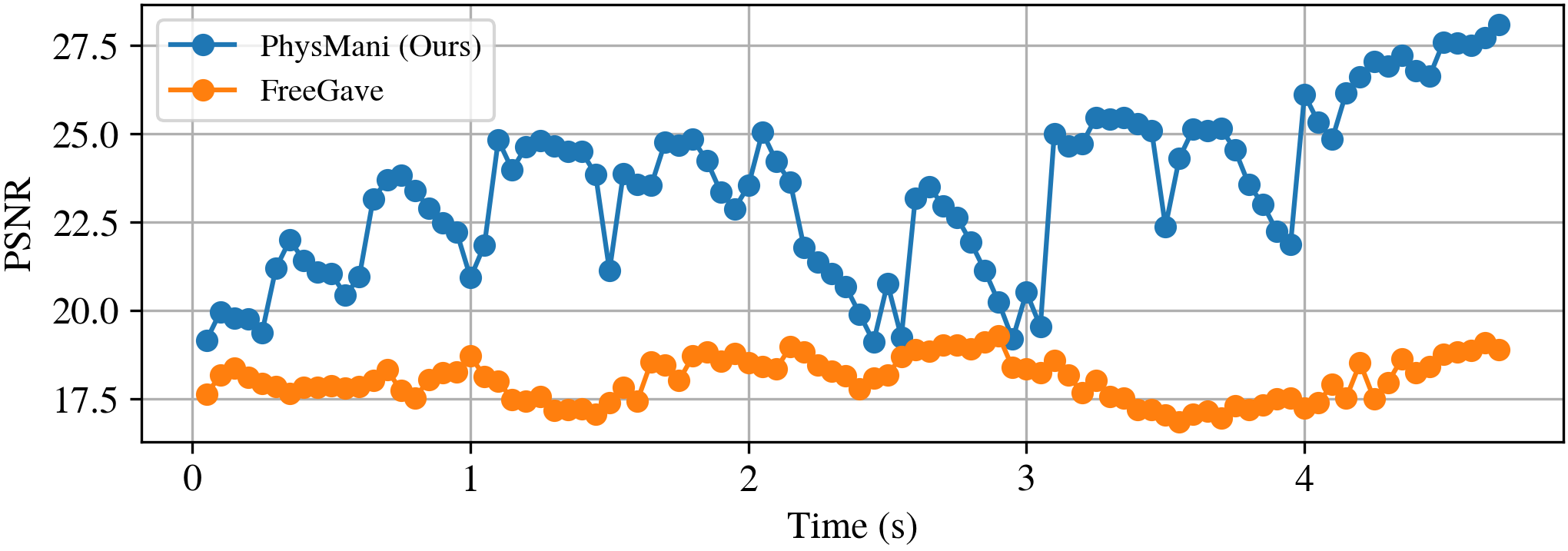}}%
    \begin{minipage}[c][\dimexpr\ht0+\dp0\relax][c]{0.31\textwidth}
        \vspace{0pt}
        \caption{The PSNR scores at each timestep, which is the mean over the next $10$ future frames. PhysMani requires 205 ms/frame, whereas FreeGave takes 607 ms/frame.}
        \label{fig:cmp_}
    \end{minipage}\hfill
    \begin{minipage}[c][\dimexpr\ht0+\dp0\relax][c]{0.66\textwidth}
        \vspace{0pt}
        \centering
        \box0
    \end{minipage}
    \par\vspace{-0.7cm}
\end{figure}

For target trajectory prediction, consecutive frames are captured at 50 ms intervals, meaning the 1st, 5th, and 10th future frames translate to horizons of 50, 250, and 500 ms. Table \ref{tab:res_future_frame} shows that \nickname{} maintains remarkably low errors of 0.008, 0.039, and 0.074 m across these timesteps. In contrast, ManiGaussian produces a substantial error of 0.388 m after just a single 50 ms step.

Figure \ref{fig:cmp_} illustrates that, for a single episode under equal compute constraints, \nickname{} runs $3.0\times$ faster and yields higher PSNR scores than FreeGave. Full test set results (Table \ref{tab:res_future_frame}) further demonstrate that under strict online time limits, our pipeline is highly efficient at predicting moving target dynamics from streaming observations compared to FreeGave. This validates the effectiveness and necessity of our modifications to FreeGave, as detailed in Section \ref{sec:phys_wm}.

\textbf{Generalization Ability:}
To further evaluate the generalization ability of our policy model pretrained on \nickname{}-Bench,
we conduct additional experiments on multiple dynamic tasks that extend our original tasks of \nickname{}-Bench by varying object shapes and colors, as well as using substantially different speeds and initial positions.
Additional results are provided in Appendix A.5. 

Overall, all these results demonstrate the physical plausibility of our future predictions and explain the higher success rate over ManiGaussian in Table~\ref{tab:res_bench}.

\subsection{Evaluation on Real-world Tasks}\label{sec:exp_real}    
\begin{table}[t]\tabcolsep= 0.25cm 
    \centering
    \caption{The success rate (SR) (\%) across all 4 real-world dynamic tasks.}\vspace{-0.3cm}
    \label{tab:real_exp}
    \resizebox{\textwidth}{!}{
    \begin{tabular}{lccccc}
    \toprule 
    & \textbf{Mean SR}
    & \textbf{Pick from Belt} 
    & \textbf{Place on Belt}
    & \textbf{Place on Rack}
    & \textbf{Remove from Rack} \\
    \midrule
    3DFA \cite{Gkanatsios2025}  & 45.3  & 75.0  & 37.5 & 6.3 &  62.5 \\
    \textbf{\nickname{}(Ours)}  & \gold{62.5}  & \gold{81.3}  & \gold{62.5} &\gold{25.0} &  \gold{81.3}  \\
    \bottomrule \vspace{0cm}
    \end{tabular}
    }
\end{table}

\begin{figure*}[t]
    \centering
    \includegraphics[width=1.0\linewidth]{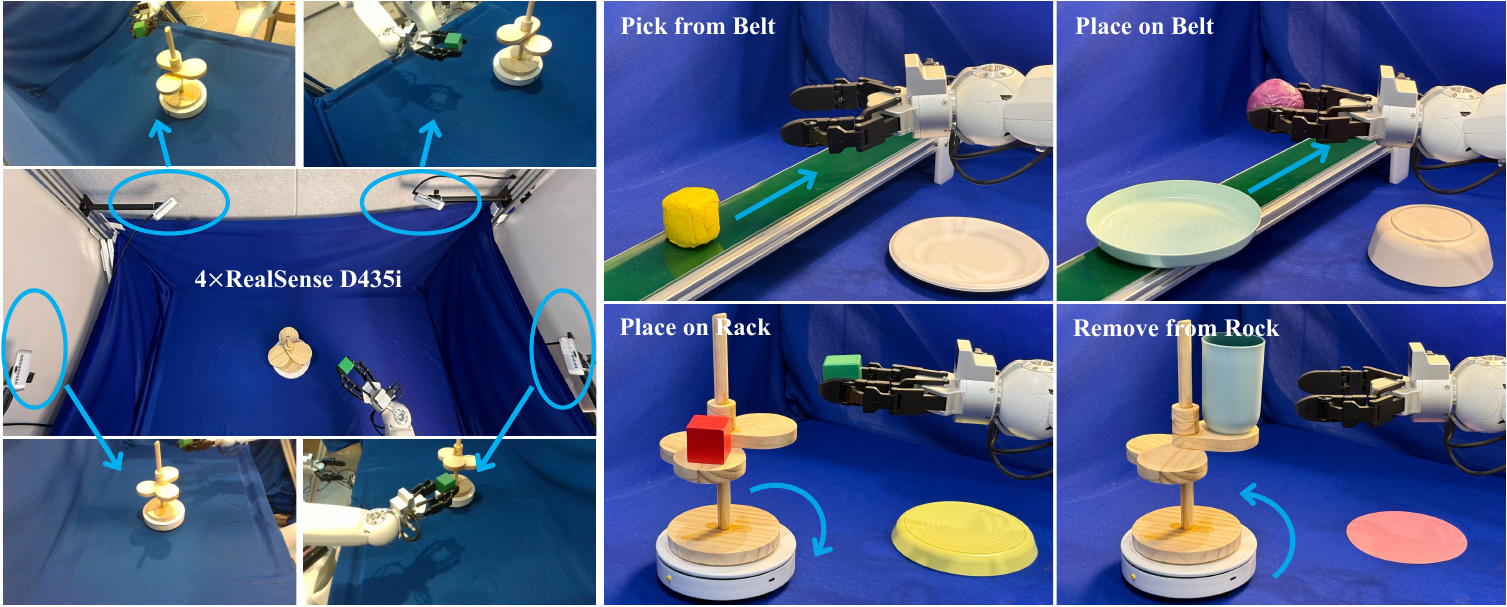}
    \vskip -0.1in
    \caption{Illustration of the physical robot setup and real-world dynamic tasks. The robot is Astribot S1\cite{gao2025towards}. \vspace{-0.3cm}}
    \label{fig:real_task_setup}
\end{figure*}
We further evaluate our approach on physical robots in four real-world dynamic tasks as described below.
Our system uses a two-fingered robotic arm in a tabletop workspace observed by four fixed RGB-D cameras (Figure \ref{fig:real_task_setup}).
All images are recorded at a resolution of $240 \times 320$, and the control frequency is set to 5\,Hz,
following common practice in recent real-world manipulation benchmarks \cite{Walke2023}.
\begin{itemize}[leftmargin=*] \vspace{-0.2cm}
\setlength{\itemsep}{1pt}
\setlength{\parsep}{1pt}
\setlength{\parskip}{1pt}
    \item \textit{picking up a moving toy}: The robot must grasp a toy such as a plastic onion and lemon on a moving belt, and then place it at a target location.
    \item \textit{placing a toy onto a moving belt}: The robot must pick up a toy such as a plastic onion and lemon, and then place it onto a moving belt. 
    \item \textit{placing a cube onto a rotating rack}: The robot must pick up a cube and then place it on a rotating rack. 
    \item \textit{removing a cup from a rotating rack}: The robot must grasp a cup from the rotating rack and place it on the table.
    \vspace{-0.2cm}
\end{itemize}

These tasks involve challenging dynamics,
with target speeds set to be comparable to the robot’s end-effector speed limits.
For each task, we collect 180 training and 60 validation episodes. We train a single model per method across all four tasks,
rather than separate models for each task. Each trained model is then evaluated online on the physical system over 64 trials,
comprising 16 varying settings such as different target speeds, target objects, and target locations, \etc{} More details of the real-world tasks and experiments are in Appendix A.6.

Table \ref{tab:real_exp} compares our method with the strongest baseline 3DFA on the four real-world dynamic tasks.
Results for the other baselines are omitted due to their rather poor performance.
Our method consistently outperforms 3DFA by a large margin, demonstrating its effectiveness in real-world dynamic manipulation.

\subsection{Ablation Study}\label{sec:ablation}

\begin{table*}[t]\tabcolsep= 0.1cm 
    \centering
    \caption{The success rate (SR) (\%) across all 16 tasks of all ablation experiments. The standard deviation is computed over two runs of the last checkpoint.}
    \label{tab:ablation}
    \resizebox{\textwidth}{!}{
    \begin{tabular}{lccccccccc}
    \toprule 
    & \multirow{2}{*}{\begin{tabular}[c]{@{}c@{}} \textbf{Mean} \\ \textbf{SR}\end{tabular}}
    & \multirow{2}{*}{\begin{tabular}[c]{@{}c@{}} \textbf{Beat} \\ \textbf{Buzz}\end{tabular}} 
    & \multirow{2}{*}{\begin{tabular}[c]{@{}c@{}} \textbf{Insert} \\ \textbf{Peg}\end{tabular}} 
    & \multirow{2}{*}{\begin{tabular}[c]{@{}c@{}} \textbf{Drop to} \\ \textbf{Hoop}\end{tabular}} 
    & \multirow{2}{*}{\begin{tabular}[c]{@{}c@{}} \textbf{Pick} \\ \textbf{Cube}\end{tabular}} 
    & \multirow{2}{*}{\begin{tabular}[c]{@{}c@{}} \textbf{Push} \\ \textbf{Button}\end{tabular}}
    & \multirow{2}{*}{\begin{tabular}[c]{@{}c@{}} \textbf{Deposit} \\ \textbf{Rubbish}\end{tabular}}
    & \multirow{2}{*}{\begin{tabular}[c]{@{}c@{}} \textbf{Place} \\ \textbf{Cup}\end{tabular}} 
    & \multirow{2}{*}{\begin{tabular}[c]{@{}c@{}} \textbf{Remove} \\ \textbf{Cup}\end{tabular}} \\
    & & & & & & & & & \\
    \midrule
    (1) removing $\bm{D}_t$ & $37.9 \pm 0.1$ & $76.2 \pm 1.2$ & $43.8 \pm 6.2$ & $30.0 \pm 2.5$ & $65.0 \pm 2.5$ & $47.5 \pm 2.5$ & \gold{$65.0 \pm 0.0$} & \gold{$21.2 \pm 1.2$} & \gold{$58.8 \pm 1.2$} \\
    (2) removing $\bm{L}$ & $40.8 \pm 1.1$ & \gold{$78.8 \pm 3.8$} & $31.2 \pm 1.2$ & $66.2 \pm 1.2$ & $72.5 \pm 0.0$ & \gold{$58.8 \pm 3.8$} & $62.5 \pm 10.0$ & $10.0 \pm 2.5$ & $58.8 \pm 8.8$ \\
    \textbf{The Full PhysMani} & \gold{$45.4 \pm 0.9$} & $76.2 \pm 3.8$ & \gold{$43.8 \pm 1.2$} & \gold{$70.0 \pm 2.5$} & \gold{$85.0 \pm 0.0$} & $53.8 \pm 1.2$ & $61.2 \pm 1.2$ & $16.2 \pm 1.2$ & $50.0 \pm 2.5$ \\
    \midrule
     
    & \multirow{3}{*}{\begin{tabular}[c]{@{}c@{}} \textit{High} \\ \textit{Speed} \\ \textbf{(H)$\xrightarrow{}$} \end{tabular}}
    & \multirow{3}{*}{\begin{tabular}[c]{@{}c@{}} \textbf{Beat} \\ \textbf{Buzz} \\ \textbf{(H)} \end{tabular}} 
    & \multirow{3}{*}{\begin{tabular}[c]{@{}c@{}} \textbf{Insert} \\ \textbf{Peg} \\ \textbf{(H)} \end{tabular}} 
    & \multirow{3}{*}{\begin{tabular}[c]{@{}c@{}} \textbf{Drop to} \\ \textbf{Hoop} \\ \textbf{(H)} \end{tabular}} 
    & \multirow{3}{*}{\begin{tabular}[c]{@{}c@{}} \textbf{Pick} \\ \textbf{Cube} \\ \textbf{(H)} \end{tabular}} 
    & \multirow{3}{*}{\begin{tabular}[c]{@{}c@{}} \textbf{Push} \\ \textbf{Button} \\ \textbf{(H)} \end{tabular}}
    & \multirow{3}{*}{\begin{tabular}[c]{@{}c@{}} \textbf{Deposit} \\ \textbf{Rubbish} \\ \textbf{(H)} \end{tabular}}
    & \multirow{3}{*}{\begin{tabular}[c]{@{}c@{}} \textbf{Place} \\ \textbf{Cup} \\ \textbf{(H)} \end{tabular}} 
    & \multirow{3}{*}{\begin{tabular}[c]{@{}c@{}} \textbf{Remove} \\ \textbf{Cup} \\ \textbf{(H)} \end{tabular}} \\
    & & & & & & & & & \\
    & & & & & & & & & \\
    \midrule
    (1) removing $\bm{D}_t$ & -- & $48.8 \pm 1.2$ & \gold{$21.2 \pm 1.2$} & $6.2 \pm 1.2$ & $18.8 \pm 6.2$ & $17.5 \pm 5.0$ & $26.2 \pm 8.8$ & \gold{$20.0 \pm 5.0$} & $40.0 \pm 0.0$ \\
    (2) removing $\bm{L}$ & -- & $42.5 \pm 2.5$ & $7.5 \pm 5.0$ & $28.8 \pm 1.2$ & $21.2 \pm 6.2$ & $18.8 \pm 6.2$ & $52.5 \pm 5.0$ & $6.2 \pm 1.2$ & $37.5 \pm 5.0$ \\
    \textbf{The Full PhysMani} & -- & \gold{$55.0 \pm 5.0$} & $20.0 \pm 7.5$ & \gold{$38.8 \pm 3.8$} & \gold{$21.2 \pm 1.2$} & \gold{$18.8 \pm 1.2$} & \gold{$53.8 \pm 1.2$} & $15.0 \pm 2.5$ & \gold{$47.5 \pm 7.5$} \\
    \bottomrule
    \end{tabular}
    }
\end{table*}

\begin{figure}[t]
    \vspace{0.2cm}
    \centering
    \setbox0\hbox{\includegraphics[width=0.70\textwidth]{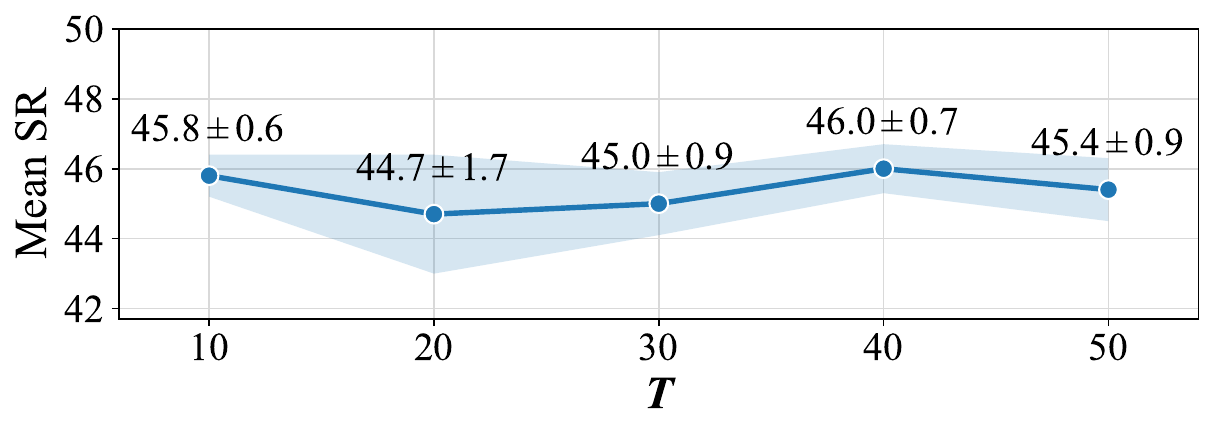}}%
    \begin{minipage}[c][\dimexpr\ht0+\dp0\relax][c]{0.26\textwidth}
        \vspace{0pt}
        \caption{Mean SR of \nickname{} over all 16 tasks under different world model optimization iterations $T$.}
        \label{fig:ablation_T_sweep}
    \end{minipage}\hfill
    \begin{minipage}[c][\dimexpr\ht0+\dp0\relax][c]{0.73\textwidth}
        \vspace{0pt}
        \centering
        \box0
    \end{minipage}
    \par\vspace{-0.7cm}
\end{figure}

Our framework consists of a physics-principled 3D world model and a future-aware policy model. 
Ablations on \nickname{}-Bench validate our design choices.

\textbf{(1) Removing the 3D scene velocity basic components $\bm{D}_t$}.
In this setting, we only keep the neighboring Gaussian coordinates for the center point $\bm{p}_t$, and feed them into the learnable token based Transformer blocks.
Basically, this is to directly evaluate the effectiveness of integrating 3D scene future dynamics. 

Note that, if we entirely remove our 3D world model and the incorporation module, the resulting framework is exactly the 3DFA backbone,
which already performs worse than our method, as shown in Tables \ref{tab:res_bench}\&\ref{tab:real_exp}.
Therefore, there is no need to conduct a separate ablation experiment.

\textbf{(2) Removing the design of learnable token $\bm{L}$}: In Step \#5 of incorporating 3D scene future dynamics (Section \ref{sec:action_policy_model}),
the learnable token $\bm{L}$ provides flexibility to fuse valuable dynamics information within a local neighborhood. To validate this design,
we remove $\bm{L}$ and instead simply adopt a self-attention mechanism to fuse predicted future dynamics within the neighborhood.  

\textbf{(3) Optimizing Gaussian velocity field with different iterations $T$}:
In our online Algorithm \ref{alg:online_opt}, we set $T=50$ to sufficiently optimize the Gaussian dynamics,
enabling more accurate future prediction. Intuitively, a smaller $T$ reduces optimization time but may degrade dynamics learning.
To evaluate this sensitivity, we conduct 4 additional ablation experiments with $T\in \{40, 30, 20, 10\}$.  

\begin{figure*}[t]
\centering
\includegraphics[width=0.96\linewidth]{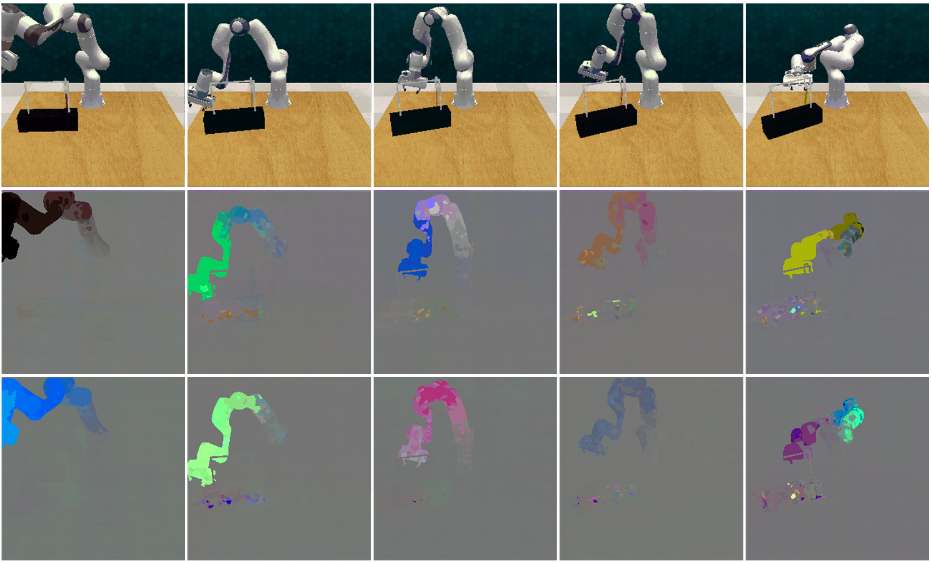}
\vspace{-0.3cm}
\caption{Visualizations of the learned six basic velocity components.}
\label{fig:vis_velocity}
\vspace{-0.5cm}
\end{figure*}

\textbf{Analysis}: Table \ref{tab:ablation} shows that:
(1) Removing the 3D scene basic velocity components $\bm{D}_t$ leads to the most significant drop in success rate, making performance comparable to the backbone 3DFA.
This demonstrates that integrating future dynamics is crucial for tackling the challenging dynamic tasks.
(2) Replacing the learnable token based mechanism with pure self-attention also noticeably degrades performance,
highlighting the effectiveness of the simple and flexible learnable token $\bm{L}$ for fusing future dynamics within a local neighborhood.
(3) Figure~\ref{fig:ablation_T_sweep} shows that our method is particularly robust to different numbers of iterations $\bm{T}$ for optimizing the velocity field, enabling our framework to quickly learn dynamics and execute actions, which is essential for dynamic manipulation. 

\vspace{-0.5cm}

\subsection{Analysis of Gaussian Velocity}

We further analyze how the learned velocity components concretely contribute to effective manipulation.
During online evaluation on \nickname{}-Bench, at a given time step t, we visualize the predicted six basic velocity components for future frames $\{10^{th}, 30^{th}, 50^{th}, 70^{th}, 90^{th}\}$, along with their corresponding rendered RGB images in Figure \ref{fig:vis_velocity}.
The second row shows the three linear velocity components, while the third row shows the three angular velocity components.
The visualizations indicate that the velocity components naturally focus on the moving robot arm and the rotating target object, while largely ignoring the static background.
This provides an explicit and physics-grounded signal that directly guides the robot’s actions.

\vspace{-0.2cm}
\subsection{Computation Efficiency}

For inference, all models are evaluated on an Intel Core i7-13700F CPU paired with an NVIDIA RTX 4090 GPU. Table~\ref{tab:efficiency} reports the total time required to train each model.
\nickname{} requires 272.8 ms to infer a single keyframe action. These keyframe actions are then converted into joint trajectories via inverse kinematics. In real-world experiments, to decouple inference from execution, we utilize an action buffer with a temporal ensemble operating at 5 Hz. Since the online optimization of the world model takes $\sim$200 ms and runs in parallel, the policy model inference remains the overall system bottleneck. Together, these design choices enable fast and dynamic manipulation.

\begin{table*}[t]\tabcolsep= 0.15cm
    \centering
    \caption{Training/inference time and memory of different methods.}\vspace{-0.3cm}
    \label{tab:efficiency}
    \resizebox{\textwidth}{!}{
    \begin{tabular}{lcccc}
        \toprule
        & \multirow{2}{*}{\begin{tabular}[c]{@{}c@{}}
            \textbf{Training Time} \\
            \textbf{(hours)}
        \end{tabular}}
        & \multirow{2}{*}{\begin{tabular}[c]{@{}c@{}}
            \textbf{Training Memory} \\
            \textbf{(GB/GPU)}
        \end{tabular}}
        & \multirow{2}{*}{\begin{tabular}[c]{@{}c@{}}
            \textbf{Inference Time} \\
            \textbf{(ms/frame)}
        \end{tabular}}
        & \multirow{2}{*}{\begin{tabular}[c]{@{}c@{}}
            \textbf{Inference Memory} \\
            \textbf{(GB)}
        \end{tabular}} \\
        & & & & \\
        \midrule
    Act3D \cite{Gervet2023}  & 19   & 18.5 & 18.8  & 1.1 \\
    3DDA \cite{Ke2024}  & 32  & 22.0 & 267.2  & 1.0 \\
    3DFA \cite{Gkanatsios2025} & 42  & 22.0 &  264.5  & 1.0 \\
    3DFA-OF      & 47  & 22.0 & 260.0  & 1.0 \\
    ManiGaussian \cite{lu2024manigaussian} & 60  & 23.5 & 897.7  & 5.9 \\
    pi0.5 \cite{Black2025}  &  149 & 15.2 & 132.7  & 7.4 \\
    \textbf{\nickname{}(Ours)}  & 53  & 22.0 &  272.8 & 1.0 \\
    \bottomrule \vspace{-0.8cm}
    \end{tabular} 
    }
\end{table*}

\section{Conclusion}\label{sec:sum}

This work presents \nickname{}, a general framework for dynamic object manipulation that explicitly models 3D scene future dynamics and conditions action prediction on them. By learning a divergence-free Gaussian velocity field online, our world model forecasts physically meaningful future dynamics with low latency, while the future-aware action policy model effectively exploits these predictions via local cross-attention. Across our newly introduced \nickname{}-Bench and challenging real-world tasks, our method consistently outperforms state-of-the-art 3D action policy and VLA-based baselines, particularly in highly dynamic tasks. These results highlight the importance of physics-principled 3D world models for embodied control and provide a practical path toward general dynamic manipulation in open environments.

\clearpage

\section*{Acknowledgments}
This work was supported in part by Research Grants Council of Hong Kong under Grants 15228626 \& 15225522 \& 15219125, and in part by Otto Poon Charitable Foundation Smart Cities Research Institute (8-CDCQ), in part by Research Center for Unmanned Autonomous Systems (1-CE3D), The Hong Kong Polytechnic University. 

%
%

\bibliographystyle{splncs04}
\bibliography{references}

\clearpage

\appendix

\section{Appendix}

\subsection{Details of Physics-principled 3D Gaussian World Model}\label{app:phys_wm}

\hspace{0.5cm}\textbf{Canonical 3D Gaussian Module}:
As shown in the left block of Figure \ref{fig:world_model}, at time $t=0$, given RGB/Ds from $C$ cameras,
denoted by $\{(I_0^c, D_0^c)\}_{c=1}^C$,  we first initialize a set of 3D Gaussian kernels $\bm{G}_0$ to represent the canonical scene geometry and appearance at time $t=0$,
based on the sparse point cloud back-projected from $C$ depth views.
We sample $25000$ points from the sparse point cloud with init scale $\bm{s}_0 =[0.04, 0.04, 0.04]$, rotation $\bm{r}_0=[0,0,0,1.0]$, opacity $\sigma=1.0$ for each Gaussian.

We train canonical 3D Gaussians for 3500 iterations and densify them after 3000 iterations. At each iteration, we sample one RGB/D view out of four camera views and optimize all Gaussian parameters using the following loss function:
\begin{equation}
\begin{split}
    \ell &= (1 - \lambda_{ssim}) \ell_1(I_0^c, \hat{I}_0^c) + \lambda_{ssim} (1 - \ell_{ssim}(I_0^c, \hat{I}_0^c)) \\
    &\quad + (1 - \lambda_{ssim}) \ell_1(D_0^c, \hat{D}_0^c) + \lambda_{ssim} (1 - \ell_{ssim}(D_0^c, \hat{D}_0^c)),
\end{split}
\label{eq:render_loss}
\end{equation}
where $\{(\hat{I}_0^c, \hat{D}_0^c)\}_{c=1}^C$ are rendered RGB/Ds from canonical 3D Gaussians and $\lambda_{ssim}=0.2$.
It takes about 15s on an RTX 4090 for \nickname{}-Bench.
\begin{figure}[ht]
    \centering
    \includegraphics[width=0.7\linewidth]{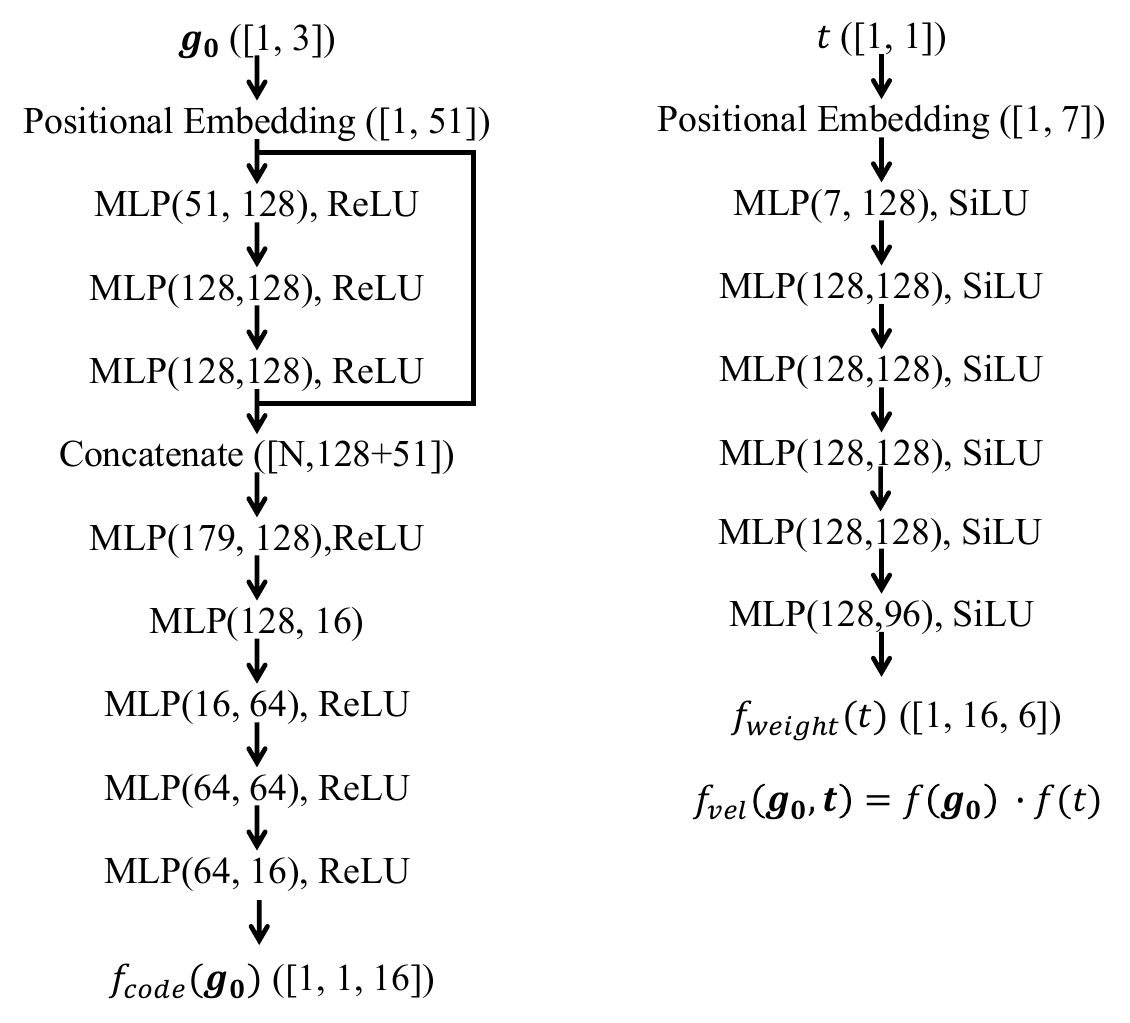}
    \caption{The network structure of $f_{vel}$.
    $\text{MLP}(51, 128)$ means a multi-layer perceptron with 51 input features and 128 hidden features.
    }
    \label{fig:fvel_structure}
\end{figure}

\textbf{Physics-principled Gaussian Velocity Module}:
The network structure of $f_{vel}$ follows FreeGave~\cite{li2025freegave}, which contains two parts: $f_{code}$ and $f_{weight}$. Figure~\ref{fig:fvel_structure} illustrates the details of $f_{vel}$. More details can be found in FreeGave.

Given new observations at current time $t$, we update $f_{vel}$ for 50 iterations (freezing all Gaussian parameters)
and then update Gaussian parameters 7 iterations (freezing $f_{vel}$), using the loss function in Equation \ref{eq:render_loss}. 

\subsection{Details of Future-aware Action Policy Model}\label{app:action_model}

\begin{figure}[t]
    \centering
    \includegraphics[width=0.5\linewidth]{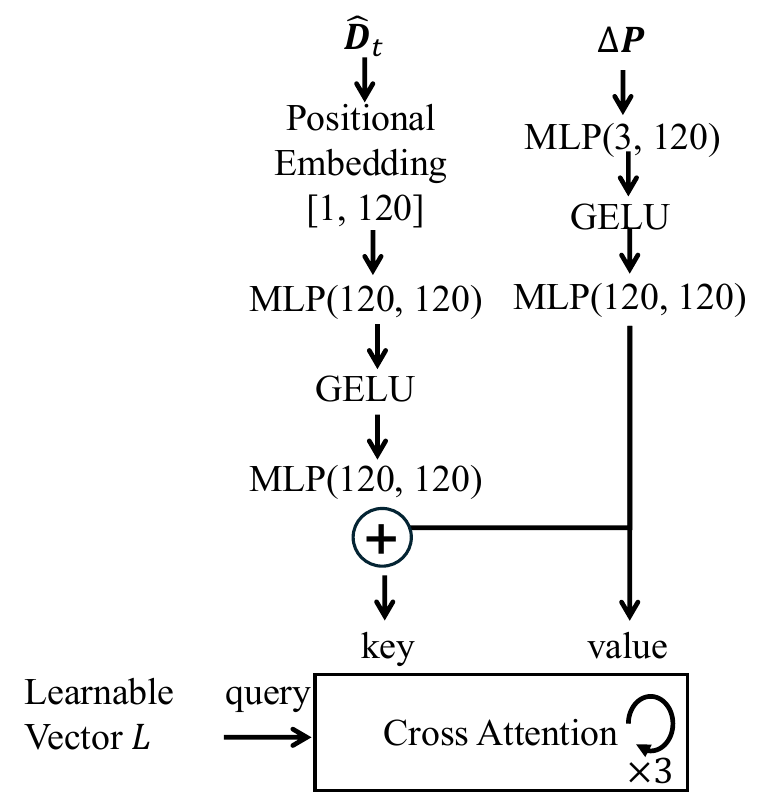}
    \caption{The network to process $\Delta \bm{P}$ and $\bm{\hat{D}}_t$ to obtain future-aware tokens $\bm{A}_t$.}
    \label{fig:appendix_policy_model}
\end{figure}

\hspace{0.5cm}\textbf{Incorporating 3D Scene Future Dynamics:}
Figure~\ref{fig:appendix_policy_model} shows the network structures processing $\Delta \bm{P}$ and $\bm{\hat{D}}_t$ to obtain future-aware tokens $\bm{A}_t$.

\textbf{Denoising End-effector's Keyposes with Rectified Flow}:
During inference, we denoise 10 steps of end-effector's keyposes with Rectified Flow, following~\cite{Gkanatsios2025}.

\textbf{Training Strategy:}
To train \nickname{}, we first pre-train the 3DFA backbone for 100,000 iterations. We then use this pre-trained model to initialize \nickname{} and fine-tune it for an additional 100,000 iterations. During fine-tuning, we employ parameter-specific learning rates: the inherited 3DFA backbone parameters are updated at one-tenth of the base learning rate ($0.1 \times lr$), while the newly introduced parameters of \nickname{} use the full base rate ($lr$). To ensure a fair comparison, the 3DFA results reported in Table~\ref{tab:res_bench} are obtained using a strictly parallel setting: 100,000 iterations of initial training followed by 100,000 iterations of fine-tuning at $0.1 \times lr$.

Specifically, \nickname{}, 3DFA, 3DDA, and Act3D are trained on 3 GPUs with a per-GPU batch size of 18, using the AdamW optimizer with a learning rate of $lr=10^{-4}$. Following its official implementation, ManiGaussian is trained on 2 RTX 3090 GPUs with a per-GPU batch size of 1. Finally, for pi0.5, we utilize an 8-GPU setup with a per-GPU batch size of 8 for a total of 400,000 iterations, employing a peak learning rate of $2.5 \times 10^{-4}$.

\subsection{Details of Implementation for Fast Computation}\label{app:fast_implementation}
We improve runtime efficiency by reusing static computation graphs (CUDA Graph optimization) during online optimization. In our setting, the number of Gaussian kernels remains is fixed across frames, which allows the update process to be executed with a static computation graph. By avoiding dynamic graph construction, we significantly reduce CUDA kernel launch overhead and improve GPU utilization. Table~\ref{tab:cudagraph_ablation} ablates the impact of static graph reuse on kernel launch time and overall update latency. Notably, incorporating CUDA Graph optimization cuts the world model latency by more than half, yielding an over $2\times$ speedup.

\begin{table}[ht]
    \centering
    \vspace{-0.4cm}
    \caption{Performance comparison with and without CUDA Graph optimization.}
    \label{tab:cudagraph_ablation}
    \begin{tabular}{lcc}
        \toprule
        \textbf{Configuration} & \textbf{\# \texttt{cudaLaunchKernel}} & \textbf{Latency (ms)} \\
        \midrule
        Without CUDA Graph optimization & 67,330 & 440 \\
        With CUDA Graph optimization (\nickname{}) & \gold{9,118} & \gold{198} \\
        \bottomrule
    \end{tabular}
    \vspace{-0.8cm}
\end{table}

\subsection{Details of \nickname{}-Bench}\label{app:bench}

\begin{figure}
    \centering
    \includegraphics[width=\linewidth, height=0.9\textheight, keepaspectratio]{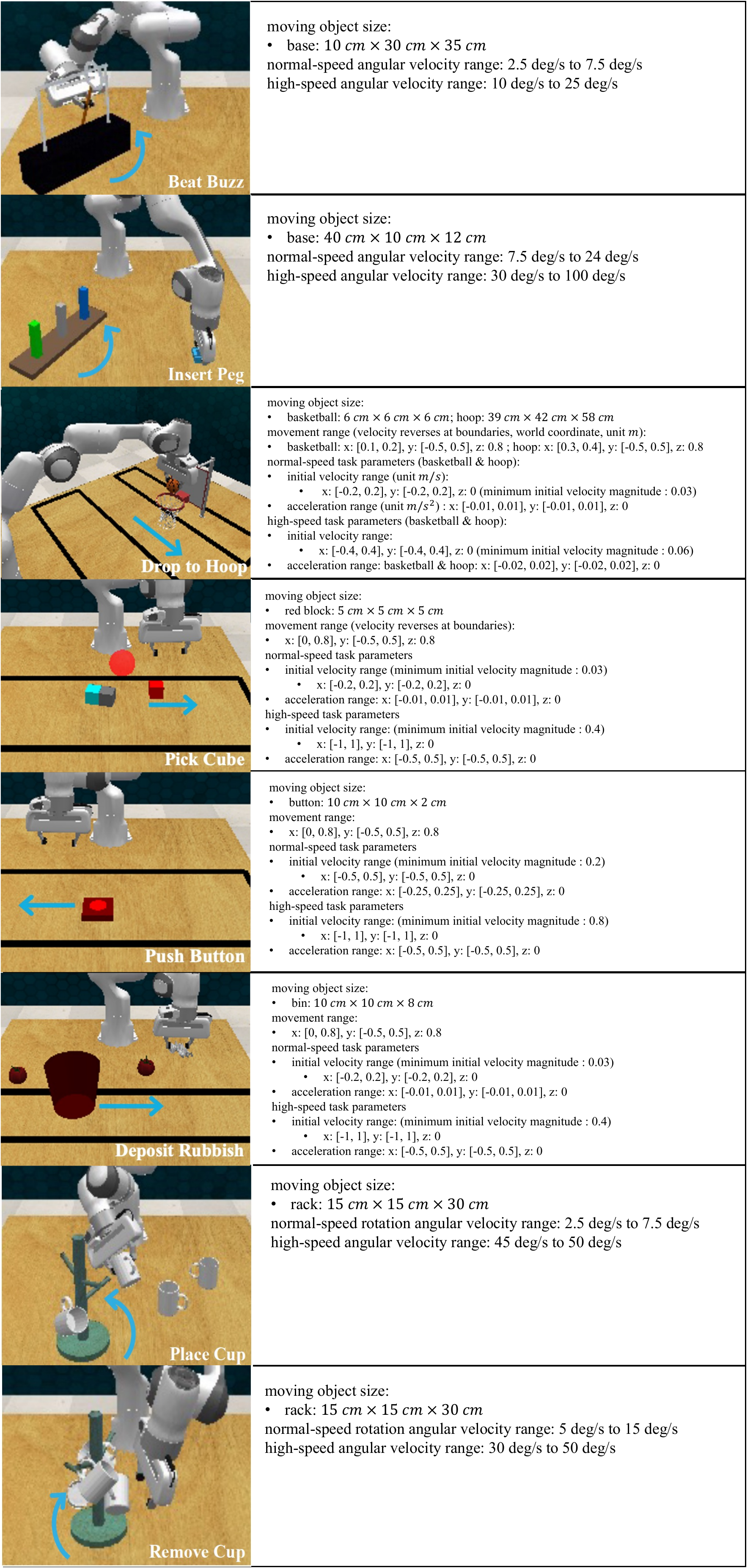}
    \caption{Detailed physical parameters of our \nickname{}-bench tasks. For example, the movement range "basketball: x: [0.0, 0.3], y: [-0.5, 0.5], z: 0.8"
    indicates that the basketball's motion is restricted to a rectangular region in the xy-plane which spans from $0.0 \sim 0.3$m along the x-axis and $-0.5\sim0.5$m along the y-axis at a fixed height of $z = 0.8$~m.}
    \label{fig:bench_details}
    \vspace{4cm}
\end{figure}

During the evaluation, we run the simulation with a time interval of 50 $ms$ and query the policy every 10 steps. To execute the predicted actions, we use RRT-connect algorithm to compute the joint trajectories. Figure \ref{fig:bench_details} summarizes the detailed physical parameters of all \nickname{}-bench tasks.

\subsection{More Results on \nickname{}-Bench}\label{app:res_bench}

\phantom{xW}\textbf{Generalization}:
To further evaluate the generalization ability of our policy model pretrained on \nickname{}-Bench, we construct novel variations of the tasks of \textit{drop basketball into moving hoop at normal/high speed}. Specifically, we independently alter the color, shape, or kinematic parameters of the moving target (\ie{}, the basketball). For each evaluation, we isolate a single attribute. For example, we apply unseen random colors, replace the spherical basketball with a cube, or randomize the initial kinematic states beyond the training distribution. During these tests, all other properties are kept strictly consistent. We directly test the pretrained policy models on these unseen configurations without any fine-tuning. As shown in Table~\ref{tab:res_bench_color_shape}, \nickname{} consistently maintains a significant performance advantage over the strong baseline 3DFA.

\begin{table*}[t]\tabcolsep=0.25cm 
    \centering
    \caption{Success rate (SR) (\%) on varying object colors, shapes, and kinematic parameters. We report the average success rate over the top 3 checkpoints by SR.}
    \label{tab:res_bench_color_shape}
    \resizebox{\textwidth}{!}{
\begin{tabular}{@{}lcc|cc|cc@{}}
\toprule
                 & \textbf{\begin{tabular}[c]{@{}c@{}}Drop to Hoop \\ (Varying Color)\end{tabular}} & \textbf{\begin{tabular}[c]{@{}c@{}}Drop to Hoop (H) \\ (Varying Color)\end{tabular}} & \textbf{\begin{tabular}[c]{@{}c@{}}Drop to Hoop \\ (Varying Shape)\end{tabular}} & \textbf{\begin{tabular}[c]{@{}c@{}}Drop to Hoop (H) \\ (Varying Shape)\end{tabular}} & \textbf{\begin{tabular}[c]{@{}c@{}}Drop to Hoop \\ (Varying Motion)\end{tabular}} & \textbf{\begin{tabular}[c]{@{}c@{}}Drop to Hoop (H) \\ (Varying Motion)\end{tabular}} \\ \hline
    3DFA & $9.2 \pm 4.2$ & $3.3 \pm 1.2$ & $22.5 \pm 7.1$ & $7.5 \pm 4.1$ & $33.3 \pm 5.9$ & $10.0 \pm 3.5$ \\
    \textbf{\nickname{}} & \gold{$24.2 \pm 3.1$} & \gold{$18.3 \pm 6.2$} & \gold{$65.0 \pm 7.4$} & \gold{$35.8 \pm 1.2$} & \gold{$66.7 \pm 2.4$} & \gold{$36.7 \pm 3.1$} \\
\bottomrule
\end{tabular}
    }\vspace{-0.1cm}
\end{table*}

\textbf{Impact of Inference Speed}:
We further evaluate our method when the policy model has different inference speeds.
We add latencies of $0/100/200/250/300$ ms to the action model on \textit{Drop Basketball into Moving Hoop}.
Our proposed \nickname{} achieves higher SR of  $77.5/57.5/37.5/35.0/20.0\%$ consistently. It outperforms 3DFA ($17.5/27.5/5.0/10.0/7.5\%$).
As the injected latency grows, SR decreases for both methods,
since the grasp tends to become unstable when actions are delayed in dynamic manipulation.
Nevertheless, \nickname{} remains stronger than 3DFA across all settings,
indicating that our 3D future-dynamics prediction helps the policy tolerate slower inference.

\subsection{Details of Real-world Dynamic Tasks and Experiments}\label{app:real_task_exp}

\hspace{0.5cm} \textbf{Robot Setup}:
We conduct real-world experiments using a 7-Degree-of-Freedom (7-DoF) two-fingered robot arm. As illustrated in Figure~\ref{fig:real_task_setup}, We mount four RealSense D435i RGBD cameras around the tabletop workspace. The cameras capture RGB images and depth data at 30 Hz with a resolution of $240\times320$. During inference, the control frequency is set to 5~Hz.

\textbf{Data collection}:
Demonstration data are collected by teleoperating the robot using a 6-DoF IMU-based controller, enabling precise and repeatable demonstrations while allowing the operator to react to dynamic targets. For each task, we collect 180 trajectories for training and 60 for validation.

\textbf{Training Details}:
The baseline 3DFA is trained for 110,000 iterations using the AdamW optimizer with a constant learning rate of $1 \times 10^{-4}$. It is trained on 3 GPUs with a per-GPU batch size of 18. To train \nickname{}, we initialize our network with the 3DFA checkpoint at 85,000 iterations and subsequently fine-tune it for another 25,000 iterations, utilizing a 4-GPU setup with a per-GPU batch size of 12. In this fine-tuning stage, we apply parameter-specific learning rates: the inherited 3DFA parameters are updated at one-tenth of the base rate ($1 \times 10^{-5}$), whereas the newly introduced parameters of \nickname{} are trained using the full base rate ($1 \times 10^{-4}$). The final model checkpoint is selected based on the minimum loss achieved on a held-out validation set.

\textbf{Evaluation}:
To comprehensively evaluate the robotic manipulation performance in dynamic scenarios,
Each task involves 4 test variables, each with 2 distinct values, yielding 16 unique test conditions. Four trials are performed under each condition to reduce the randomness of single-run experiments. A trial is considered successful only if all following criteria are satisfied: 
\begin{itemize}[leftmargin=*] \vspace{-0.2cm}
\setlength{\itemsep}{1pt}
\setlength{\parsep}{1pt}
\setlength{\parskip}{1pt}
    \item The robot stably grasps the target object without dropping or misgrasping. 
    \item The object is placed into the target region without displacement or tipping. 
    \item No severe collisions occur between the robot and the moving belt, rotating rack, target area, or other components.
    \item The process is completed within a predefined reasonable time limit (60s). 
\end{itemize}

\hspace{0.2cm}\textbf{Task Descriptions}:
Here we detail the four dynamic tasks and their specific variable settings:

\textit{Picking Up a Moving Toy:} Robot grasps a moving toy (\eg{}, a plastic onion or lemon) from a conveyor belt and places it into a designated stationary dish. To evaluate the robustness of the policy, we introduce variations across four dimensions: the initial position of the toy on the belt, the type of toy (onion or lemon), the location of the target dish on the tabletop, and the speed of the conveyor belt (normal or high).

\textit{Placing a Toy Onto a Moving Belt:} Robot picks up a stationary toy from the tabletop and accurately places it into a dish moving uniformly along a conveyor belt. The task conditions vary based on four parameters: the initial position of the moving dish on the belt (two values covering different workspace regions), the type of toy (onion or lemon), the initial tabletop location of the toy, and the conveyor belt speed (normal or high).

\textit{Placing a Cube onto a Rotating Rack:} Robot grasps a cube and deposits it onto a continuously rotating rack. We diversify the experimental setups by altering four variables: the initial angular position of the rotating frame, the color of the grasped cube (red or green), the direction of rotation (clockwise or counterclockwise), and the rotational speed of the rack (normal or high).

\textit{Removing a Cup from a Rotating Rack:} Robot tracks and grasps a moving cup from the rotating rack and transfers it to a designated target area. The variations for this task mirror the dynamic parameters of the previous task, including two initial angular positions for the rotating frame, two cup colors (blue or green), two rotation directions (clockwise or counterclockwise), and two rotational speeds.

\end{document}